\documentclass{article}
\pdfoutput=1
\usepackage{ijcai23}

\usepackage{times}
\usepackage{soul}
\usepackage{url}
\usepackage[utf8]{inputenc}
\usepackage[small]{caption}
\usepackage{graphicx}
\usepackage{amsmath}
\usepackage{amsthm}
\usepackage{booktabs}
\usepackage{algorithm}
\usepackage{algorithmic}
\usepackage[switch]{lineno}

\usepackage{amssymb}
\usepackage[T1]{fontenc}
\usepackage{multirow}
\usepackage{forloop}
\usepackage{pgfplots}
\usepackage{rotating}
\usepackage{subcaption}
\usepackage{tikz}
\usepackage{hyperref}

\hypersetup{hidelinks}
\pgfplotsset{compat=1.17}
\usetikzlibrary{shapes.misc, positioning, shapes, shadows}

\title{Disentanglement of Latent Representations via Causal Interventions}

\author{
Ga\"{e}l Gendron
\and
Michael Witbrock\and
Gillian Dobbie\\
\affiliations
University of Auckland
}

\DeclareMathOperator*{\argmax}{argmax}

\begin{document}

\maketitle

\begin{abstract}

The process of generating data such as images is controlled by independent and unknown factors of variation. The retrieval of these variables has been studied extensively in the disentanglement, causal representation learning, and independent component analysis fields. Recently, approaches merging these domains together have shown great success. 
Instead of directly representing the factors of variation, the problem of disentanglement can be seen as finding the interventions on one image that yield a change to a single factor.
Following this assumption, we introduce a new method for disentanglement inspired by causal dynamics that combines causality theory with vector-quantized variational autoencoders. Our model considers the quantized vectors as causal variables and links them in a causal graph. 
It performs causal interventions on the graph and generates atomic transitions affecting a unique factor of variation in the image. 
We also introduce a new task of action retrieval that consists of finding the action responsible for the transition between two images. 
We test our method on standard synthetic and real-world disentanglement datasets. We show that it can effectively disentangle the factors of variation and perform precise interventions on high-level semantic attributes of an image without affecting its quality, even with imbalanced data distributions.

\end{abstract}

\section{Introduction}

The problem of recovering the mechanisms underlying data generation, particularly for images, is challenging and has been widely studied in machine learning research \cite{DBLP:conf/icml/LocatelloBLRGSB19,DBLP:conf/nips/GreseleKSSB21,DBLP:journals/pieee/ScholkopfLBKKGB21,DBLP:conf/clear2/LachapelleRSEPL22}. The disentanglement field aims to represent images as high-level latent representations where such mechanisms, or \textit{factors of variation}, are divided into separate, e.g. orthogonal, dimensions \cite{DBLP:journals/corr/abs-1812-02230}.
By contrast, causal representation learning attempts to recover such factors as causal variables sparsely linked in a graph \cite{DBLP:journals/pieee/ScholkopfLBKKGB21,DBLP:journals/corr/abs-2209-11924}. Despite the similarities between the two problems, until recently little work has attempted to combine the two fields \cite{DBLP:conf/icml/SuterMSB19,DBLP:conf/cvpr/YangLCSHW21}. Some approaches have also borrowed ideas from independent component analysis \cite{DBLP:conf/nips/GreseleKSSB21,DBLP:conf/clear2/LachapelleRSEPL22}.
A central concept linking this work is the Independent Causal Mechanisms (ICM) principle \cite{DBLP:journals/pieee/ScholkopfLBKKGB21} which states that the generative process of a data distribution is made of independent and autonomous modules.
In order to recover these modules, disentanglement approaches mainly rely on variational inference and Variational Auto-Encoders (VAEs) \cite{DBLP:conf/icml/LocatelloBLRGSB19} or Generative Adversarial Networks 
\cite{DBLP:conf/nips/GabbayCH21}.
Despite the success of vector-quantized VAE architectures for generating high-quality images at scale \cite{DBLP:conf/nips/OordVK17,DBLP:conf/nips/RazaviOV19,DBLP:conf/icml/RameshPGGVRCS21}, they have not been considered in the disentanglement literature, except in the speech synthesis domain \cite{DBLP:conf/icassp/WilliamsZCY21}.

In this paper, we attempt to bridge this gap by proposing a novel way to represent the factors of variation in an image using quantization. We introduce a Causal Transition (CT) layer able to represent the latent codes generated by a quantized architecture within a causal graph and allowing causal interventions on the graph. We consider the problem of disentanglement as equivalent to recovering the atomic transitions between two images $X$ and $Y$. In this setting, one high-level action causes an intervention on the latent space, which generates an atomic transition. This transition affects only one factor of variation. 
We use our architecture for two tasks: given an image, act on one factor of variation and generate the intervened-on output; and given an input-output pair, recover the factor of variation whose modification accounts for the difference. 
To study the level of disentanglement of latent quantized vectors, we also introduce a Multi-Codebook Quantized VAE (MCQ-VAE) architecture, dividing the VQ-VAE latent codes into several vocabularies. Figure \ref{fig:ct_vae} illustrates our full architecture. We show that our model can effectively disentangle the factors of variation in an image and allow precise interventions on a single factor without affecting the quality of the image, even when the distribution of the factors is imbalanced.

We summarise our contributions as follows:
(i) We introduce a novel quantized variational autoencoder architecture and a causal transition layer.
(ii) We develop a method to perform atomic interventions on a single factor of variation in an image and disentangle a quantized latent space, even with imbalanced data. 
(iii) Our model can learn the causal structure linking changes on a high-level global semantic concept to low-level local dependencies.
(iv) We propose a new task of recovering the action that caused the transition from an input to an output image.
(v) We show that our model can generate images with and without interventions without affecting quality.
Our code and data are available here: \href{https://github.com/Strong-AI-Lab/ct-vae}{https://github.com/Strong-AI-Lab/ct-vae}.

\section{Related Work}

\paragraph{Disentanglement}

There is no formal definition of disentanglement, but it is commonly described as the problem of extracting the \textit{factors of variation} responsible for data generation \cite{DBLP:conf/icml/LocatelloBLRGSB19}. These factors of variation are usually considered independent variables associated with a semantic meaning \cite{DBLP:conf/icml/MathieuRST19,DBLP:journals/pieee/ScholkopfLBKKGB21}. Formally, it amounts to finding, for an image $X$, the factors $Z = \{Z_i\}_{i \in 1 \dots D}$ s.t. $f(Z_1, \dots, Z_D) = X$, $D$ being the dimensionality of the latent space. Modifying the value of one $Z_i$ modifies a single semantic property of $X$ (e.g. the shape, the lighting, or the pose) without affecting the other properties associated with the values $Z_{j \neq i}$. The main disentanglement methods are based on the regularisation of Variational Autoencoders (VAEs) \cite{DBLP:journals/corr/KingmaW13}. Unsupervised models comprise the $\beta$-VAE \cite{DBLP:conf/iclr/HigginsMPBGBML17}, the $\beta$-TCVAE \cite{DBLP:conf/nips/ChenLGD18}, the FactorVAE \cite{DBLP:conf/icml/KimM18} or the DIP-VAE \cite{DBLP:conf/iclr/0001SB18}. However, unsupervised approaches have been challenged, and the claim that fully unsupervised disentanglement is achievable remains under debate \cite{DBLP:conf/icml/LocatelloBLRGSB19,DBLP:conf/nips/HoranRW21}. More recent approaches rely on weak supervision 
\cite{DBLP:conf/icml/LocatelloPRSBT20,DBLP:conf/nips/GabbayCH21}. 
Our approach belongs to this category. In particular, the CausalVAE \cite{DBLP:conf/cvpr/YangLCSHW21} generates a causal graph to link the factors of variation together. Our approach also attempts to take advantage of causal models for disentanglement, but the two methods differ greatly. We consider the problem of disentanglement from the perspective of causal dynamics and use quantization instead of a standard VAE to generate the causal variables.

\paragraph{Quantization}

The Vector-Quantized VAE (VQ-VAE) \cite{DBLP:conf/nips/OordVK17} is an autoencoder where the encoder generates a discrete latent vector instead of a continuous vector $Z \in \mathbb{R}^D$. From an input image, the VQ-VAE builds a discrete latent space $\mathbb{R}^{K \times D}$ with $K$ vectors representing the quantization of the space. As an analogy, these vectors can be interpreted as words in a codebook of size $K$. The encoder samples $N \times N$ vectors from the latent space when building $Z \in \mathbb{R}^{N \times N \times D}$. Each sampled vector describes the local information in a $N \times N$ grid representing an abstraction of the input image $X$. The VQ-VAE and its derivations have proven very successful at generating high-quality images at scale \cite{DBLP:conf/nips/RazaviOV19,DBLP:conf/icml/RameshPGGVRCS21}. The Discrete Key-Value Bottleneck \cite{DBLP:journals/corr/abs-2207-11240} builds upon the VQ-VAE architecture, introducing a key-value mechanism to retrieve quantized vectors and using multiple codebooks instead of a single one; the method is applied to domain-transfer tasks. To the best of our knowledge, we are the first to apply quantized autoencoders to disentanglement problems.

\paragraph{End-to-end Causal Inference}

Causal tasks can be divided into two categories: causal structure discovery and causal inference. Causal structure discovery consists in learning the causal relationships between a set of variables with a Direct Acyclic Graph (DAG) structure, while causal inference aims to estimate the values of the variables \cite{pearl2009causality} quantitatively.
In our work, we attempt to recover the causal structure responsible for the transition from an input image $X$ to an input image $Y$ and perform causal inference on it to retrieve the values of the missing variables. As the causal graph acts on latent variables, we also need to retrieve the causal variables, i.e. the disentangled factors of variation, $Z$.

The Structural Causal Model (SCM) \cite{pearl2009causality} is a DAG structure representing causal relationships on which causal inference can be performed. Causal queries are divided into three layers in Pearl's Causal Hierarchy (PCH) \cite{bareinboim2022pearl}: associational, interventional and counterfactual. Our work attempts to solve interventional queries, i.e. questions of the type "how would  $Y$ evolve if we modify the value of $X$?", represented by the formula $P(Y=y|\mathbf{do}(X=x))$. The \textbf{do}-operation \cite{pearl2009causality} corresponds to the attribution of the value $x$ to the variable $X$ regardless of its distribution. The Causal Hierarchy Theorem (CHT) \cite{bareinboim2022pearl} states that interventional data is necessary to solve interventional queries. Accordingly, the data we use is obtained by performing interventions $a$ on images $X$.

Recent work performed causal structure discovery and inference in an end-to-end fashion, like VCN \cite{DBLP:journals/corr/abs-2106-07635} and DECI \cite{DBLP:journals/corr/abs-2202-02195}. Our approach is similar, as we want to identify and estimate the causal links end-to-end. The main differences are that we do not assume linear relationships as in VCN, and the causal variables are unknown in our problem and must also be estimated.
This problem of retrieving causal variables is known as causal representation learning \cite{DBLP:journals/pieee/ScholkopfLBKKGB21}. In particular, our method is close to interventional causal representation learning \cite{DBLP:journals/corr/abs-2209-11924}.

\paragraph{Graph Neural Networks}

Graph Neural Networks are a family of Deep Learning architectures operating on graph structures. A graph $\mathcal{G} = \langle V, E \rangle$ is a set of nodes $V$ and edges $E$ where an edge $e_{ij} \in E$ links two nodes $v_i$ and $v_j$. A feature vector $x_i \in \mathcal{D}$ is associated with each node $v_i$. A feature matrix $X \in \mathbb{R}^{|V| \times D}$ represents the set of vectors. The graph is represented with an adjacency matrix $A \in [0,1]^{|V| \times |V|}$. Graph neural networks aggregate the node features based on the neighbourhood of each node. A generic representation is shown in Equation \ref{eq:gnn}.

\begin{equation}
    X^{(l+1)} = GNN(X^{(l)}; A)
    \label{eq:gnn}
\end{equation}

The most popular GNN architectures are GCN \cite{DBLP:conf/iclr/KipfW17}, GAT \cite{DBLP:conf/iclr/VelickovicCCRLB18}, 
and GraphSAGE \cite{DBLP:conf/nips/HamiltonYL17}. Recently, Graph Neural Networks have proved themselves a suitable architecture for causal inference tasks 
\cite{DBLP:journals/corr/abs-2109-04173}
because of their ability to represent interventional queries on a causal graph. iVGAE \cite{DBLP:journals/corr/abs-2109-04173} and VACA \cite{DBLP:conf/aaai/Sanchez-MartinR22} are variational autoencoders operating on graph structures and able to perform \textbf{do}-operations. Our work differs from these models as they assume the causal graph structure to be known, whereas it is unknown in our problem.

\paragraph{Causal Dynamics}

World models 
attempt to learn a latent representation capturing the dynamics of an environment over time. The Variational Causal Dynamics (VCD) model \cite{lei2022causal} learns the invariant causal mechanisms responsible for the evolution of an environment under the influence of an action $a$. In such systems, the environment transits from the state $(s)$ to $(s+1)$ because of $a$. We approach disentanglement from this angle, considering our representation disentangled if, when applying an action $a_i$, we intervene on a single factor of variation $Z_i$. The state $(s)$ corresponds to the input image $X$, and the state $(s+1)$ corresponds to an output image $Y$ after intervention $a$ on one factor of variation of $X$. The main difference between our work and VCD is that our environment is latent and must be discovered.

\section{Causal Transition Variational Autoencoder}

\subsection{Problem Definition}

The problem we aim to solve can be divided into two parts. The first is a disentanglement problem; we aim to generate a disentangled latent representation to which we can apply an intervention on a specific feature, e.g. change the pose, the lighting or the shape of the object shown. The second problem is reverse-engineering the intervention responsible for the transition from an input to an output image: given the input image and the output image after an intervention, identify the action that caused the transition.

We name the input image $X$ and the output image after transition $Y$, with $a$ being the cause of the change. Given a set $\mathcal{S}$ of pairs of input-output images $(X_1, Y_1), (X_2, Y_2), \dots \in \mathcal{S}$ s.t. $\forall (X,Y) \in \mathcal{S}, Y=f_a(X)$, we aim to find the function $f_a$. The first problem is returning $Y$ given $X$ and $a$, and the second problem is extracting $a$ given $X$ and $Y$. The causal queries associated are $P(Y|X,do(a))$ and $P(a|X,Y)$.

\subsection{Overview of Our Method}

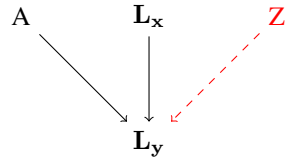
\begin{figure}
    \centering
    \begin{tikzpicture}[node distance=1.7cm]

        \node (a) {A};
        \node (x) [right of=a] {$\mathbf{L_x}$};
        \node (z) [right of=x] {\textcolor{red}{Z}};
        \node (y) [below of=x] {$\mathbf{L_y}$};
        
        \draw[->,dashed,red] (z) -- (y);
        \draw[->] (a) -- (y);
        \draw[->] (x) -- (y);     
        
    \end{tikzpicture}
    \caption{Causal Graph of the transition in latent space. $\mathbf{L_x}$ is the latent representation of the input image $X$, and $\mathbf{L_y}$ is the latent representation of the output image $Y$. The transition from $X$ to $Y$ depends on the representations of $X$ and $Y$ and the actions causing the transition. These actions are divided between labelled actions A, which can be controlled, and unknown actions Z, represented by a stochastic process, typically $Z \sim \mathcal{N}(0,1)$. }
    \label{fig:latent_causal_grah}
\end{figure}

We generate disentangled latent representations $\mathbf{L_x}$ and $\mathbf{L_y}$, and apply a causal transition model on them to represent the function $f_a$. The corresponding causal graph is illustrated in Figure \ref{fig:latent_causal_grah}. 

\begin{figure*}[t]
    \centering
    \begin{subfigure}{0.45\textwidth}
        \centering
        \includegraphics[width=\linewidth]{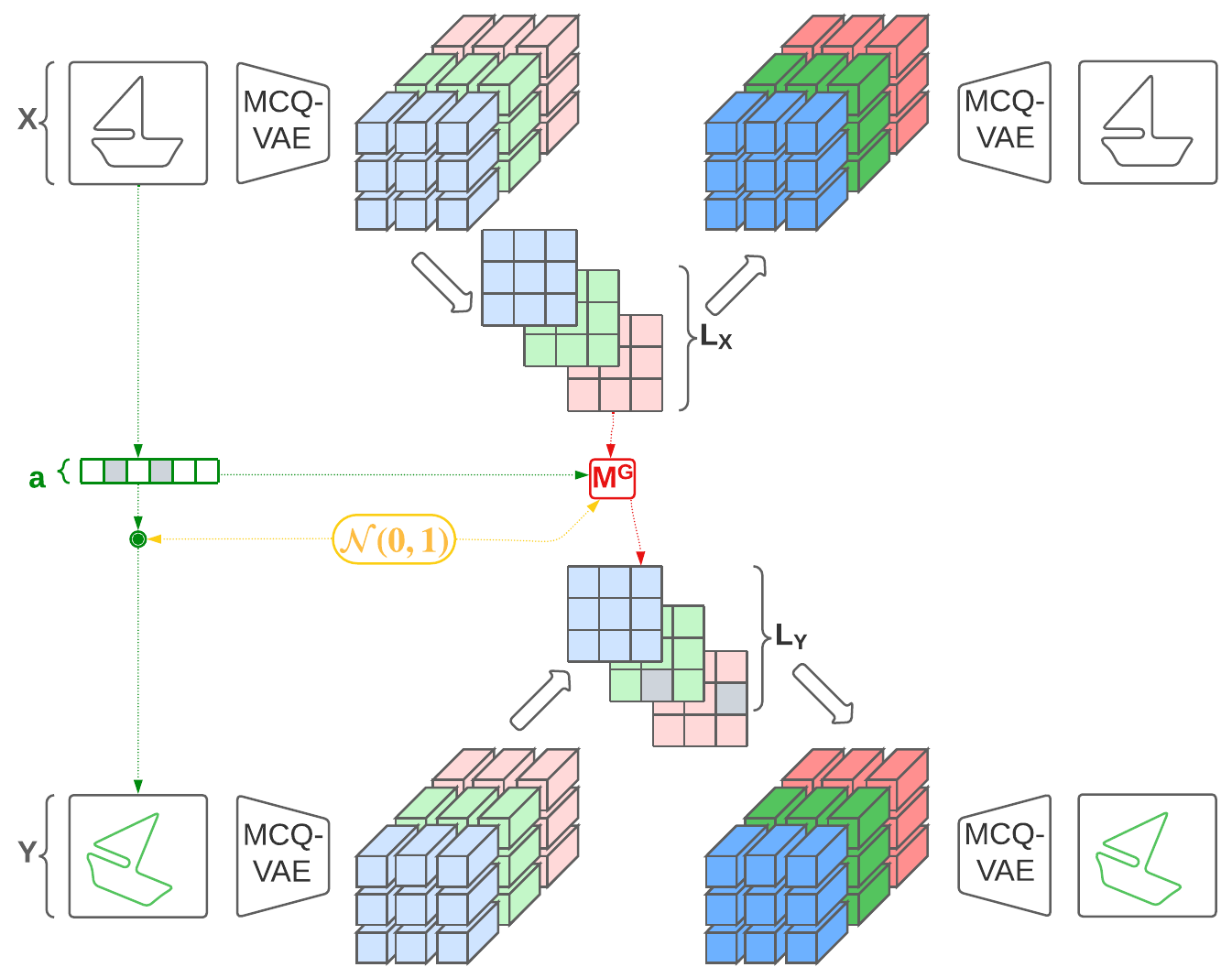}
    \caption{Overview of the CT-VAE with three codebooks. }
    \label{fig:ct_vae_overview}
    \end{subfigure}
    \hfill
    \begin{subfigure}{0.52\textwidth}
        \centering
        \begin{subfigure}{0.37\textwidth}
            \includegraphics[width=\linewidth,trim=0 0.1cm 0 0.1cm,clip]{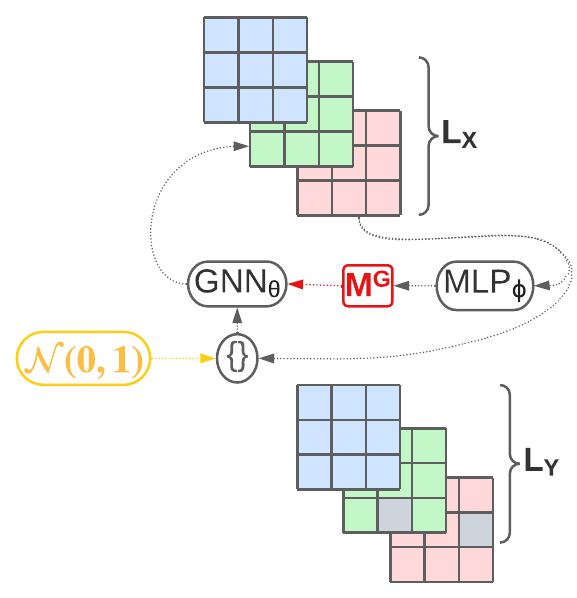}
            \vspace{-0.5cm}
            \caption{Standard mode. }
            \label{fig:ct_vae_standard}
        \end{subfigure}
        \hfill
        \begin{subfigure}{0.58\textwidth}
            \includegraphics[width=\linewidth,trim=0 0.1cm 0 0.1cm,clip]{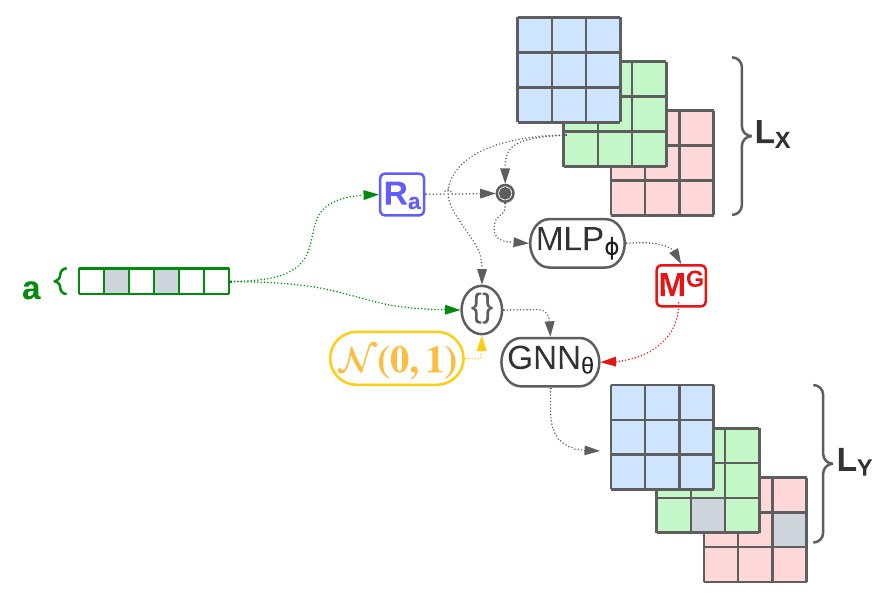}
            \vspace{-0.5cm}
            \caption{Action mode. }
            \label{fig:ct_vae_action}
        \end{subfigure}
        \hfill
        \begin{subfigure}{0.66\textwidth}
            \includegraphics[width=\linewidth,trim=0 0.1cm 0 0.1cm,clip]{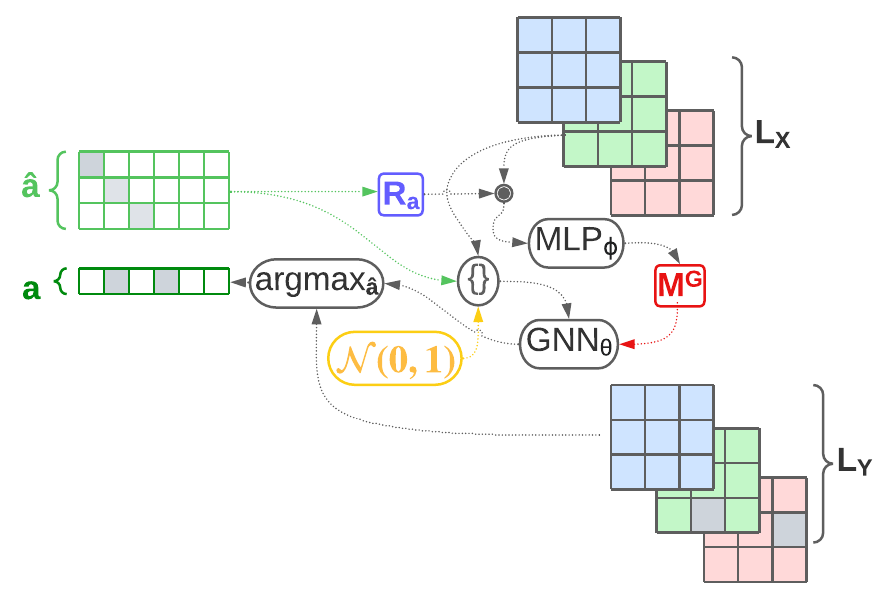}
            \vspace{-0.5cm}
            \caption{Causal mode. }
            \label{fig:ct_vae_causal}
        \end{subfigure}
    \end{subfigure}
    \caption{CT-VAE architecture and the three modes of inference. The model is trained to encode and decode an image under an intervention $\mathbf{a}$. The MCQ-VAE generates a quantized latent space and the CT layer performs causal reasoning on that space to modify it according to the intervention. A masked MLP generates the causal graph from the quantized codes under intervention and a GNN infers the corresponding output latent quantized codes from it. In \textit{standard} mode, the CT layer attempts to reproduce the initial space $\mathbf{L_x}$. In \textit{action} mode, it attempts to transpose $\mathbf{L_x}$ to the latent space of the output image $\mathbf{L_y}$. The \textit{causal} mode consists in retrieving the intervention responsible for a transition between $X$ and $Y$. The action maximising the likelihood of $\mathbf{L_y}$ is selected. }
    \label{fig:ct_vae}
\end{figure*}

We use an autoencoder to encode the input image into a latent space $\mathbf{L_x}$ and then decode it. We use a VQ-VAE for this autoencoder; more details are given in Section \ref{sec:mcq-vae}.
We then build a causal model of the transition from $\mathbf{L_x}$ to $\mathbf{L_y}$. This model attempts to learn two components: a vector $\mathbf{a}$ representing the action taken to transition from $\mathbf{L_x}$ to $\mathbf{L_y}$ and an adjacency matrix $\mathbf{M^\mathcal{G}}$ representing the causal dependencies in the latent space with respect to this action (the dependencies are not the same, for example,  if the action affects the position \textit{vs} the colour of the image). $\mathbf{M^\mathcal{G}}$ is specific to an instance but $\mathbf{a}$ models the invariant causal generative mechanisms of the task. In other words, $\mathbf{a}$ represents the \textit{why}, and $\mathbf{M^\mathcal{G}}$ represents the \textit{what} and \textit{how}. A comprehensive description is provided in Section \ref{sec:latent_causal_model}. Figure \ref{fig:ct_vae_overview} shows an overview of our model.

\subsection{Multi-Codebook Quantized VAE}
\label{sec:mcq-vae}

We introduce a new autoencoder architecture based on the VQ-VAE called \textit{Multi-Codebook Quantized VAE} or MCQ-VAE. As in \cite{DBLP:journals/corr/abs-2207-11240}, our model allows the latent space to have multiple codebooks to increase the expressivity of the quantized vectors. As shown in Figure \ref{fig:ct_vae_overview}, each vector is divided into several sub-vectors belonging to a different codebook. In the VQ-VAE, each latent vector embeds local information, e.g. the vector on the top-left corner contains the information needed to reconstruct the top-left corner of the image. Using multiple codebooks allows us to disentangle the local representation into several modules that can be reused across the latent vectors and combined. Each sub-vector is linked to one causal variable in the causal transition model. The downside of this division into codebooks is memory consumption, which increases linearly with the number of codebooks.

\subsection{Latent Causal Transition Model}
\label{sec:latent_causal_model}

The autoencoder described in the previous section generates the latent space suitable for our causal transition algorithm. The algorithm can be used in two ways. To apply a transformation on the latent space corresponding to an action with a high-level semantic meaning. Alternatively, given the result of the transformation, to retrieve the corresponding action. To address these two goals and the global reconstruction objective of the autoencoder, we divide our method into three operating modes:

\begin{itemize}
    \item \textit{Standard:} illustrated in Figure \ref{fig:ct_vae_standard}, there is no causal transition, this mode reconstructs the input image $X$,
    \item \textit{Action:} illustrated in Figure \ref{fig:ct_vae_action}, a causal transition is applied given an action, the autoencoder must return the image after transition $Y$,
    \item \textit{Causal:} illustrated in Figure \ref{fig:ct_vae_causal}, given the two images $X$ and $Y$, before and after transition, the algorithm returns the corresponding action.
\end{itemize}

\paragraph{Causal Inference}

The transition from $\mathbf{L_x}$ to $\mathbf{L_y}$ is made using a Graph Neural Network (GNN) architecture where $[\mathbf{L_x}, \mathbf{a}, \mathbf{z}]$ are the nodes and $\mathbf{M^\mathcal{G}}$ is the adjacency matrix. 

\begin{equation}
    \mathbf{L_y} = GNN_\theta([\mathbf{L_x}, \mathbf{a}, \mathbf{z}];\mathbf{M^\mathcal{G}})
\end{equation}

Therefore, the transition problem can be translated to a node classification task. For each variable $L_{xi} \in \mathbf{L_x}$, we aim to find the corresponding $L_{yi}$ based on its parents $pa(L_{yi})$. As shown in Figure \ref{fig:latent_causal_grah}, the parents of $L_{yi}$ are the action $\mathbf{a}$, a subset of the variables in $\mathbf{L_x}$, and some exogenous variables that are unknown and modelled by a probability distribution $\mathbf{z} \sim \mathcal{N}(0,1)$. 
The action $\mathbf{a}$ has a global impact that may depend on the node position. To take this into account, we add a positional embedding to each node.
The choice of GNNs for the architecture is motivated by their permutation-equivariance property. The second motivation is their ability to model causal graphs where the variables are multi-dimensional \cite{DBLP:journals/corr/abs-2109-04173}.

\paragraph{Causal Structure Discovery}

The causal graph $\mathcal{G}$ is represented by a learned adjacency matrix $\mathbf{M^\mathcal{G}}$. As in previous works \cite{DBLP:journals/corr/abs-2106-07635,lei2022causal}, the coefficients $\alpha_{ij}$ of $\mathbf{M^\mathcal{G}}$ are obtained using Bernoulli trials with parameters determined by a dense network. 

\begin{equation}
    \begin{aligned}
        \{M^\mathcal{G}_{ij}\} &\sim \text{Bernouilli}(\sigma(\alpha_{ij})) \\
        \alpha_{ij} &= MLP_\phi([L_{xi},L_{xj}]; \mathbf{a})
    \end{aligned}
    \label{eq:structure_discovery}
\end{equation}

\begin{figure}
    \centering
    \begin{tikzpicture}[node distance=2cm]

        \node (lx) {$[L_{xi}, \{L_{xj}\}_j]$};
        \node[rectangle,draw] (mlp1) [right of=lx] {$MLP^{a_1}_\phi$};
        \node[rectangle,draw,double] (mlp2) [below of=mlp1, node distance=0.8cm] {$MLP^{a_2}_\phi$};
        \node[rectangle,draw,double] (mlp0) [above of=mlp1, node distance=0.8cm] {$MLP^{\emptyset~}_\phi$};
        
        \node (a) [below of=lx] {$a_2$};
        \node (at) [above of=a, node distance=0.7cm] {};
        \node[rectangle,draw] (ra) [right of=a] {$\mathbf{R}^{\mathcal{A}}_{a_2,L_{xi}}$};

        \node[circle,draw] (dot0) [right of=mlp0] {\textbullet};
        \node[circle,draw] (dot2) [right of=mlp2, node distance=1.4cm] {\textbullet};
        \node[circle,draw] (p) [right of=dot0, node distance=1.4cm] {+};
        
        \node (alpha) [right of=p, node distance=1.4cm] {$\mathbf{\alpha_i}$};
        
        \draw[->] (a) -- (ra);
        \draw[-] (a) -- (at.center);
        \draw[->] (at.center) -| (mlp2.south);
        \draw[->] (lx) -- (mlp0.west);
        \draw[->] (lx) -- (mlp2.west);
        \draw[->] (mlp0) -- (dot0);
        \draw[->] (mlp2) -- (dot2);
        \draw[->] (ra) -| (dot0);
        \draw[->] (ra) -| (dot2);
        \draw[->] (dot0) -- (p);
        \draw[->] (dot2) -| (p);
        \draw[->] (p) -- (alpha);
        
    \end{tikzpicture}
    \caption{Structure of the causal discovery model in \textit{action} mode. The probability vector $\mathbf{\alpha_i}$ of dependencies of $L_{xi}$ is computed by a dense network with inputs the variable $L_{xi}$ and every other variable $L_{xj}$. The action $a$ determines which intervention network to use and the mask $\mathbf{R}^{\mathcal{A}}_{a,L_{xi}}$ selects either the intervention network or the network corresponding to no intervention $\emptyset$. 
    }
    \label{fig:causal_masking}
\end{figure}
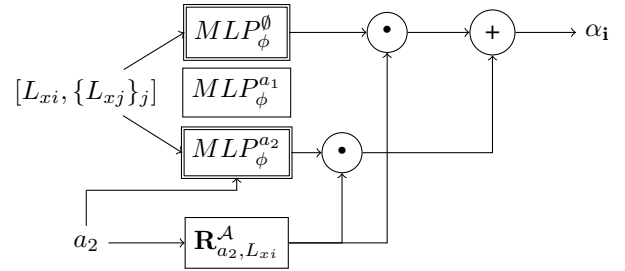

$\sigma(\cdot)$ is an activation function.
We use separate networks for each intervention following the Independent Causal Mechanism (ICM) principle \cite{DBLP:journals/pieee/ScholkopfLBKKGB21}, which states that the mechanisms involved in data generation do not influence each other. 
As in \cite{lei2022causal}, we use an intervention mask $\mathbf{R}^{\mathcal{A}}$ to determine which network to use, with $\mathcal{A}$ the set of possible actions $\mathbf{a} \in \mathcal{A}$. Each network computes the probability of existence of a link in the causal graph between two variables $L_{xi}$ and $L_{xj}$ under an intervention from $\mathcal{A}$. $\mathbf{R}_\mathbf{a}$ is a binary vector determining whether each causal variable is affected by action $\mathbf{a}$ or not (action $\emptyset$) and selecting the appropriate sub-network as shown on Equation \ref{eq:intervention_mask} and Figure \ref{fig:causal_masking}.

\begin{equation}
    \begin{aligned}
        MLP_\phi([L_{xi},L_{xj}]; \mathbf{a}) &= (MLP^{\mathbf{a}}_\phi([L_{xi},L_{xj}]))^{\mathbf{R}^{\mathcal{A}}_{a,L_{xi}}} \\
        &\cdot (MLP^{\emptyset}_\phi([L_{xi},L_{xj}]))^{1 - \mathbf{R}^{\mathcal{A}}_{a,L_{xi}}}
    \end{aligned}
    \label{eq:intervention_mask}
\end{equation}

We require $|\mathcal{A}|+1$ networks as we can have $|\mathcal{A}|$ possible interventions, or no intervention, on a given variable. The intervention mask $\mathbf{R}^{\mathcal{A}}$ is jointly learned with $\phi$.

\paragraph{Causal Attribution}

Finally, in \textit{causal} mode the action $\mathbf{a}$ is obtained by selecting from the set $\mathcal{A}$ of possible actions the one corresponding to the optimal transition to $\mathbf{L_y}$.

\begin{equation}
    \begin{aligned}
        &\mathbf{a} = \argmax\limits_{\hat{\mathbf{a}} \in \mathcal{A}}\left( \mathbb{E}_{\mathbf{L_y}}[GNN_\theta([\mathbf{L_x}, \hat{\mathbf{a}}, \mathbf{z}];\mathbf{M^\mathcal{G}})] \right) \\
        &\text{with } \mathbf{M^\mathcal{G}} \sim \text{Bernouilli}(\sigma(MLP_\phi([\mathbf{L_x}]; \hat{\mathbf{a}})))
    \end{aligned}
    \label{eq:causal_a}
\end{equation}

\subsection{Training}

The model is trained in two stages. First, we pre-train the MCQ-VAE on a reconstruction task using the same procedure as in the original VQ-VAE. Second, we plug the Causal Transition layer into the architecture and train it on the transition task. The weights of the MCQ-VAE are frozen during this stage. Several losses and regularisation methods are added to help the model perform causal transition. During this stage, the learning process is divided further into three alternating steps. These steps correspond to the \textit{standard}, \textit{action}, and \textit{causal} modes.

\paragraph{Standard} In \textit{standard} mode, the transition model must behave like the identity function, as shown in Figure \ref{fig:ct_vae_standard}. Given $\mathbf{L_x}$ and the \textit{null} action $\emptyset$, the causal graph $\mathbf{M^\mathcal{G}}$ should be the identity matrix $\mathbf{I}$ and $\mathbf{L_y}$ equals $\mathbf{L_x}$.

\begin{equation}
    \begin{aligned}
        \mathcal{L}_x(\phi,\theta) &= \mathbb{E}_{\mathbf{L_x}}[GNN_\theta([\mathbf{L_x},\mathbf{z}]; \mathbf{M^\mathcal{G}})] \\
        \text{with } \mathbf{M^\mathcal{G}} &\sim \text{Bernoulli}(\sigma(MLP_\phi([\mathbf{L_x}];\emptyset)))
    \end{aligned}
    \label{eq:loss_x}
\end{equation}

The primary loss function used is represented on Equation \ref{eq:loss_x}. It maximizes the likelihood of the generated representation, driving it towards $\mathbf{L_x}$.
In addition to this loss function, two regularisation losses are used.

\begin{equation}
    \mathcal{L}_{id_y}(\theta) = \mathbb{E}_{\mathbf{L_x}}[GNN_\theta([\mathbf{L_x},\mathbf{z}];\mathbf{I})]
    \label{eq:loss_idy}
\end{equation}

\begin{equation}
    \mathcal{L}_{id_{M^\mathcal{G}}}(\phi) = \lVert \text{Bernoulli}(\sigma(MLP_\phi([\mathbf{L_x}];\emptyset)))) - \mathbf{I} \rVert^2
    \label{eq:loss_idm}
\end{equation}

The loss function in Equation \ref{eq:loss_idy} maximizes the likelihood of the output of the GNN parameterised by $\theta$ given a causal graph being equal to the identity, and the one in Equation \ref{eq:loss_idm} regularises $\mathbf{M^\mathcal{G}}$ and the parameters $\phi$ towards the identity matrix. As in \cite{lei2022causal}, the Straight-Through Gumbel-Softmax reparametrisation trick \cite{DBLP:conf/iclr/JangGP17} is used to allow the gradient to flow through the Bernoulli sampling. The last two losses are only used in \textit{base} mode.

\paragraph{Action} In \textit{action} mode, the transition model must transform $\mathbf{L_x}$ into $\mathbf{L_y}$, as shown in Figure \ref{fig:ct_vae_action}. This is a two-steps process. First, given $\mathbf{L_x}$ and $\mathbf{a}$, the model learns $\mathbf{M^\mathcal{G}}$. Second,  given $\mathbf{L_x}$, $\mathbf{a}$, and $\mathbf{M^\mathcal{G}}$, the model infers $\mathbf{L_y}$.

\begin{equation}
    \begin{aligned}
        \mathcal{L}_y(\phi,\theta) &= \mathbb{E}_{\mathbf{L_y}}[GNN_\theta([\mathbf{L_x},\mathbf{a},\mathbf{z}]; \mathbf{M^\mathcal{G}})] \\
        \text{with } \mathbf{M^\mathcal{G}} &\sim \text{Bernoulli}(\sigma(MLP_\phi([\mathbf{L_x}];\mathbf{a})))
    \end{aligned}
    \label{eq:loss_y}
\end{equation}

The loss function in Equation \ref{eq:loss_y} ensures that the transition model parameterized by $\theta$ and $\phi$ accurately generates $\mathbf{L_y}$ in \textit{action} mode. The Straight-Through Gumbel-Softmax \cite{DBLP:conf/iclr/JangGP17} reparametrisation trick is used again. This loss function is identical to the first one introduced in \textit{standard} mode, but given an intervention.

\paragraph{Causal} In \textit{causal} mode, the model does not output a reconstructed image but an action vector, as shown in Figure \ref{fig:ct_vae_causal}. The decoder is not called, instead we introduce a loss function maximising the likelihood of the generated action vector.

\begin{equation}
    \begin{aligned}
        \mathcal{L}_\mathbf{a}(\phi,\theta) &= \mathbb{E}_\mathbf{a}[\mathbf{q}]\\
        &\text{with } q_{\hat{\mathbf{a}}} = \frac{e^{\mathbb{E}_{\mathbf{L_y}}[GNN_\theta([\mathbf{L_x}, \hat{\mathbf{a}}, \mathbf{z}];\mathbf{M^\mathcal{G}}(\hat{\mathbf{a}}))]}}{\sum\limits_{\mathbf{a} \in \mathcal{A}} e^{\mathbb{E}_{\mathbf{L_y}}[GNN_\theta([\mathbf{L_x}, \mathbf{a}, \mathbf{z}];\mathbf{M^\mathcal{G}}(\mathbf{a}))]}} \\
        &\text{and } \mathbf{M^\mathcal{G}}(\mathbf{a}) \sim \text{Bernouilli}(\sigma(MLP_\phi([\mathbf{L_x}]; \mathbf{a})))
    \end{aligned}
    \label{eq:loss_a}
\end{equation}

The output in \textit{causal} mode is a vector $\mathbf{q} \in \mathbb{R}^{|\mathcal{A}|}$ corresponding to the probability for each action to be the cause of the transition from $\mathbf{L_x}$ to $\mathbf{L_y}$. It is obtained by computing the transition in \textit{action} mode for each action in $\mathcal{A}$ and its likelihood given the true $\mathbf{L_y}$. The likelihoods are converted to probabilities using softmax activation. The resulting vector $\mathbf{q}$ is trained to resemble the true action vector $\mathbf{a}$ using the loss $\mathcal{L}_\mathbf{a}(\phi,\theta)$.

\paragraph{Graph Regularisation}

The likelihood of the output $\mathbf{L_y}$ cannot be directly optimised because the causal graph $\mathcal{G}$ is unknown and acts as a latent variable. In consequence, we maximise the Evidence Lower Bound (ELBO) shown in Equation \ref{eq:elbo}, as in VCN and VCD \cite{DBLP:journals/corr/abs-2106-07635,lei2022causal}.

\begin{equation}
    \mathbb{ELBO}(\phi, \theta) = \mathbb{E}_{q_{\phi}(\mathcal{G})}[p_{\theta}(\mathbf{L_y}|\mathcal{G})] - \text{KL}(q_{\phi}(\mathcal{G}) || p(\mathcal{G}))
    \label{eq:elbo}
\end{equation}

The first term is the reconstruction term corresponding to the losses $\mathcal{L}_x$ and $\mathcal{L}_y$ introduced previously in this section. We now derive the regularisation term $\text{KL}(q_{\phi}(\mathcal{G})|| p(\mathcal{G}))$ where KL is the Kullback–Leibler divergence, $p(\mathcal{G})$ is the prior distribution over causal graphs, and $q_{\phi}(\mathcal{G})$ is the learned posterior distribution, as shown on Equations \ref{eq:structure_discovery} and \ref{eq:elbo_posterior}.

\begin{equation}
            q_{\phi}(M^{\mathcal{G}}_{ij}) = q_{\phi}(\alpha_{ij}|\mathbf{a}, L_{xi}, L_{xj})\\
            = \sigma(\alpha_{ij})
    \label{eq:elbo_posterior}
\end{equation}

Unlike in VCN, we do not need to constrain the space of graphs to Directed Acyclic Graphs (DAGs), as our graph is a DAG by construction. All edges start in the set $\mathbf{L_x}$ and end in the set $\mathbf{L_y}$. Thus, the prior probability of the existence of an edge follows the uniform law:

\begin{equation}
    p(\mathbf{M}^{\mathcal{G}}_{ij}) 
    = \text{Uniform}(0,1)
    \label{eq:elbo_prior}
\end{equation}

To help regularise the causal graph $\mathcal{G}$ further, two other losses are introduced for the set of parameters $\phi$.

\begin{equation}
    \mathcal{L}_{|M^\mathcal{G}
    |}(\phi) = \lVert \text{Bernoulli}(\sigma(MLP_\phi([\mathbf{L_x}];\mathbf{a}))) \rVert^2
    \label{eq:loss_size_g}
\end{equation}

Equation \ref{eq:loss_size_g} reduces the norm of the generated causal graph and, by extension, minimises the number of dependencies of the causal variables.

\begin{equation}
    \mathcal{L}_{dep(M^\mathcal{G})
    }(\phi) = \sum\limits_i \lVert \prod\limits_j (1 - \sigma(MLP_\phi([L_{xi}, L_{xj}];\mathbf{a}))) \rVert^2
    \label{eq:loss_posg}
\end{equation}

Finally, Equation \ref{eq:loss_posg} minimises, for each node, the joint probability of having no dependencies, and ensures that at least one dependency will exist for every node of the graph.

\section{Experiments}

\subsection{Datasets}

We perform our experiments on several standard disentanglement benchmarks. The Cars3D dataset \cite{DBLP:conf/nips/ReedZZL15} contains 3D CAD models of cars with 3 factors of variation: the type of the car, camera elevation, and azimuth. The Shapes3D dataset \cite{DBLP:conf/icml/KimM18} contains generated scenes representing an object standing on the floor in the middle of a room with four walls. The scene contains 6 factors of variation: the floor, wall and object colours, the scale and shape of the object, and the orientation of the camera in the room. The Sprites dataset \cite{DBLP:conf/nips/ReedZZL15} contains images of animated characters. There are 9 variant factors corresponding to character attributes such as hair or garments. The DSprites dataset \cite{DBLP:conf/iclr/HigginsMPBGBML17} contains 2D sprites generated based on 6 factors: the colour, shape, and scale of the sprite, the location of the sprite with x and y coordinates, and the rotation of the sprite. 

All the datasets described above are synthetic, and all of their generative factors of variation are labelled. We also apply our model to real-world data. The CelebA dataset \cite{DBLP:conf/iccv/LiuLWT15} is a set of celebrity faces labelled with 40 attributes including gender, hair colour, and age. Unlike the above datasets, these attributes do not fully characterise each image. Many attributes linked to the morphology of the face are not captured or are captured with insufficient precision to uniquely correspond to an individual. These missing attributes correspond to exogenous factors of variation.

We build the transitions $(X, Y)$ using the given factors of variation. For instance, two images $X$ and $Y$ can be part of a transition if all their factors are identical but one. We generate the transitions $(X,Y)$ and $(Y,X)$ with two opposite actions $a$ and $-a$ updating the value of the corresponding factor of variation.

\paragraph{Imbalanced Data}

\begin{figure}
    \centering
    \scriptsize
    \begin{tikzpicture}
        \begin{axis}[
                width=8cm, 
                height=4.5cm,
                ybar stacked,
                ylabel = {Distribution},
                symbolic x coords={Cars3D, Shapes3D, Sprites, DSprites},
                xtick=data,
                cycle list = {green!60,blue!40,red!30,yellow!80}
                ]
            \addplot+[fill] coordinates {(Cars3D,0.02) (Shapes3D,0.18) (Sprites,0.13) (DSprites,0.03)};
            \addplot+[fill] coordinates {(Cars3D,0.11) (Shapes3D,0.18) (Sprites,0.13) (DSprites,0.05)};
            \addplot+[fill] coordinates {(Cars3D,0.87) (Shapes3D,0.18) (Sprites,0.19) (DSprites,0.35)};
            \addplot+[fill] coordinates {(Cars3D,0) (Shapes3D,0.14) (Sprites,0.09) (DSprites,0.28)};
            \addplot+[fill] coordinates {(Cars3D,0) (Shapes3D,0.07) (Sprites,0.06) (DSprites,0.28)};
            \addplot+[fill] coordinates {(Cars3D,0) (Shapes3D,0.26) (Sprites,0.13) (DSprites,0)};
            \addplot+[fill] coordinates {(Cars3D,0) (Shapes3D,0) (Sprites,0.09) (DSprites,0)};
            \addplot+[fill] coordinates {(Cars3D,0) (Shapes3D,0) (Sprites,0.06) (DSprites,0)};
            \addplot+[fill] coordinates {(Cars3D,0) (Shapes3D,0) (Sprites,0.11) (DSprites,0)};
        \end{axis} 
    \end{tikzpicture}
    \caption{Distribution of the factors of variation for each dataset. The longer the bar the higher the number of variations for the corresponding factor. }
    \label{fig:data_imbalance}
\end{figure}
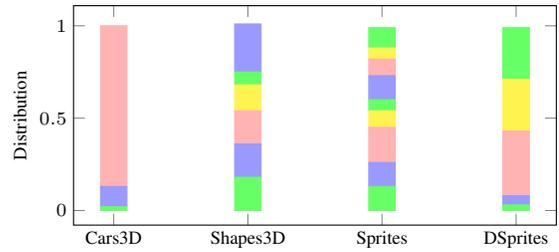

Figure \ref{fig:data_imbalance} shows the distribution of actions in the datasets. The factors are highly unbalanced for every dataset. For instance, the Cars3D dataset has three factors of variation. The first one (in \textcolor{green!60}{green}) has few variations in the distribution, the second (in \textcolor{blue!40}{blue}) has six times more variations, and the third one (in \textcolor{red!30}{red}) has thirty times more variations. The data are not i.i.d. To tackle this issue, we adopt a model-centric approach powered by causality theory: the causal graph built by our model aims to eliminate the effect of the spurious correlations induced by the data imbalance. Our experiments show that our model can learn to distinguish the factors of variation efficiently, and significantly reduces the effect of confounders.

\subsection{Image Generation Under Intervention}
\label{sec:exp_image_gen}

\begin{figure}[t]
    \centering
    \scriptsize
    \newcounter{actionnum}
    \newcommand\nb{4}
    \newcommand\nnb{5}
    \newcommand{\testfolder}{test1}
    \begin{tabular}{cccccc}
         & Input 
        \forloop{actionnum}{1}{\value{actionnum} < \nnb}{ & Output \arabic{actionnum}} \\
        \begin{sideways} Floor hue \end{sideways} & \includegraphics[width=0.12\linewidth,trim=20 20 20 20,clip]{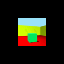}
        \forloop{actionnum}{0}{\value{actionnum} < \nb}{
            & \includegraphics[width=0.12\linewidth,trim=20 20 20 20,clip]{results/\testfolder/CT-VAE-V5_output_action_3dshapes_floor_hue_+\arabic{actionnum}_\testfolder.png}
        } \\
        \begin{sideways} Wall hue \end{sideways} & \includegraphics[width=0.12\linewidth,trim=20 20 20 20,clip]{results/\testfolder/CT-VAE-V5_input_action_3dshapes_\testfolder.png}
        \forloop{actionnum}{0}{\value{actionnum} < \nb}{
            & \includegraphics[width=0.12\linewidth,trim=20 20 20 20,clip]{results/\testfolder/CT-VAE-V5_output_action_3dshapes_wall_hue_+\arabic{actionnum}_\testfolder.png}
        } \\
        \begin{sideways} Object hue \end{sideways} & \includegraphics[width=0.12\linewidth,trim=20 20 20 20,clip]{results/\testfolder/CT-VAE-V5_input_action_3dshapes_\testfolder.png}
        \forloop{actionnum}{0}{\value{actionnum} < \nb}{
            & \includegraphics[width=0.12\linewidth,trim=20 20 20 20,clip]{results/\testfolder/CT-VAE-V5_output_action_3dshapes_object_hue_+\arabic{actionnum}_\testfolder.png}
        } \\
        \begin{sideways} Scale \end{sideways} & \includegraphics[width=0.12\linewidth,trim=20 20 20 20,clip]{results/\testfolder/CT-VAE-V5_input_action_3dshapes_\testfolder.png}
        \forloop{actionnum}{0}{\value{actionnum} < \nb}{
            & \includegraphics[width=0.12\linewidth,trim=20 20 20 20,clip]{results/\testfolder/CT-VAE-V5_output_action_3dshapes_scale_+\arabic{actionnum}_\testfolder.png}
        } \\
        \begin{sideways} Shape \end{sideways} & \includegraphics[width=0.12\linewidth,trim=20 20 20 20,clip]{results/\testfolder/CT-VAE-V5_input_action_3dshapes_\testfolder.png}
        \forloop{actionnum}{0}{\value{actionnum} < \nb}{
            & \includegraphics[width=0.12\linewidth,trim=20 20 20 20,clip]{results/\testfolder/CT-VAE-V5_output_action_3dshapes_shape_+\arabic{actionnum}_\testfolder.png}
        } \\
        \begin{sideways} Orientation \end{sideways} & \includegraphics[width=0.12\linewidth,trim=20 20 20 20,clip]{results/\testfolder/CT-VAE-V5_input_action_3dshapes_\testfolder.png}
        \forloop{actionnum}{0}{\value{actionnum} < \nb}{
            & \includegraphics[width=0.12\linewidth,trim=20 20 20 20,clip]{results/\testfolder/CT-VAE-V5_output_action_3dshapes_orientation_+\arabic{actionnum}_\testfolder.png}
        } \\
    \end{tabular}
    \caption{Atomic interventions on one factor of variation on images from the Shapes3D dataset. Each row corresponds to an intervention on a different factor with the first row being the input image. Output $i$ corresponds to the output after applying the same action $i$ times. }
    \label{fig:images_shapes3D}
\end{figure}

We perform a series of interventions on input images and study the quality of the generated images. After each intervention, we take the result of generation and apply a new intervention to it. Figure \ref{fig:images_shapes3D} illustrates the result for the Shapes3D dataset. We can observe that the reconstructed images do not undergo a quality reduction. This is expected as our method does not affect the codebooks created by vector quantization. We can also see that, after intervention, the reconstructed images have only the intervened-on factor modified. For example, background colours are not modified when operating on the object colour. Similarly, the more complex intervention on the camera orientation involves many changes in the pixels of the image but is correctly handled by the CT-VAE. 
Therefore, our method can properly disentangle the factors of variation and discriminate among the variables affected and unaffected by the intervention.
We can be observe a few exceptions. Changing the shape of the object generates slight modifications of the background near the object. As we use the output for the next generation, these modifications may propagate. 
Further studies and results for the other datasets are given in the appendix.

\subsection{Causal Structure Discovery}

\begin{figure}[t]
    \centering
    \scriptsize
    \begin{tabular}{ccccc}
        & Cars3D & Shapes3D & Sprites & DSprites \\
        \begin{sideways} Adjacency \end{sideways} &
         \includegraphics[width=0.16\linewidth]{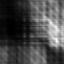} &
         \includegraphics[width=0.16\linewidth]{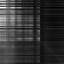} & \includegraphics[width=0.16\linewidth]{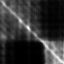} & \includegraphics[width=0.16\linewidth]{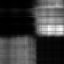} \\
        \begin{sideways} Mask \end{sideways} & 
        \includegraphics[width=0.16\linewidth]{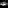} &
        \includegraphics[width=0.16\linewidth]{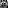} & \includegraphics[width=0.16\linewidth]{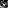} & \includegraphics[width=0.16\linewidth]{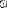} \\
    \end{tabular}
    \caption{Causal structure of the CT layer for each dataset. Masks are $H \times W$ matrices with $H$ and $W$ respectively the height and width of the latent image. Adjacency matrices are $HW \times HW$ matrices. Adjacencies and masks are averaged over a batch. The brightness indicates the probability of existence of a link between two variables for the adjacency matrix. For the mask, it indicates the probability that the variable undergoes an intervention. }
    \label{fig:causal_discovery_results}
\end{figure}

We now look at the structure of our causal transition model. Figure \ref{fig:causal_discovery_results} shows the generated latent adjacency matrices and the causal masks. The dependencies are very different depending on the dataset on which the model was trained. In the Cars3D dataset, the variables mainly look at the bottom half of the latent image. The nature of the car images can explain this behaviour; all the Cars3D cars are located in the middle, on a uniform background. The car can be slightly below the middle, depending on the camera orientation. Thus, the elements of the image affected by an intervention  are also in the middle of the image. This behaviour is consistent with the masked image, which shows that the zones where the intervention takes place match the possible locations of the car. This behaviour is not observed for the Shapes3D dataset, where the background is also affected by interventions.

\paragraph{Action Recovery} As detailed in Section \ref{sec:exp_image_gen}, the CT-VAE supports interventions affecting a single factor of variation. Given the causal graph, we would like to see whether this factor can be recovered. A factor of variation has a value evolving along an axis, either increasing or decreasing. We represent actions as one-hot vectors, so increments and decrements are considered different actions. We consider the problem of recovering the factor of variation, and the problem of recovering the action, i.e. the factor of variation and the direction. Table \ref{tab:causal_results} summarises the results. The CT-VAE can retrieve with high accuracy the correct action for the Cars3D and Shapes3D datasets but struggle with Sprites and DSprites, which contain smaller sprites than the former datasets. For the Sprites dataset, the model has trouble identifying the direction of the action but can retrieve the correct factor in most cases. We can observe that the number of actions has little impact on the accuracy.

\begin{table}[t]
    \centering
    \begin{tabular}{lrrrr}
        \hline
         & Cars3D & Shapes3D & Sprites & DSprites \\
        \hline
        \#Actions & 6 & 12 & 18 & 10 \\
        Action Acc. & 0.71 & 0.83 & 0.42 & 0.44 \\
        Factor Acc. & 0.94 & 0.90 & 0.82 & 0.58 \\
        \hline
    \end{tabular}
    \caption{Accuracy of \textit{causal} mode. The task aims to retrieve the action causing the transition from image $X$ to $Y$. \textit{\#Actions} shows the number of possible action, i.e. the cardinality of the output space. Each factor of variation has two associated actions, one increasing and the other decreasing its value. \textit{Action Acc.} shows the action accuracy and \textit{Factor Acc.} shows the factor accuracy.  }
    \label{tab:causal_results}
\end{table}

\section{Discussion and Conclusion}

Recovering the mechanisms generating images is a challenging task. Current disentanglement methods rely on Variational Auto-Encoders and attempt to represent the various factors of variation responsible for data generation on separate dimensions. We propose a new method based on causality theory to perform disentanglement on quantized VAEs. Our method can perform interventions on the latent space affecting a single factor of variation. We test it on synthetic and real-world data.

A limitation of our current architecture is the division between the pre-training and fine-tuning stages. Codebooks are fixed in the fine-tuning stage, limiting the CT layer in both the level of expressivity and the disentanglement of latent codes. Training  the two parts of the model jointly on a reconstruction and transition task could alleviate this issue but would require regularising the distribution of latent codes. Our model is also limited by the set of actions, which must be known in advance. In future work, we will attempt to solve these issues, including learning the set of possible actions.

One of the questions that our method raises regards the level of disentanglement of the latent space. The latent space studied in this paper is of a very different nature from the ones studied in the standard disentanglement literature. The VAE latent space is traditionally a $\mathbb{R}^D$ vector where each dimension accounts for one factor of variation if accurately disentangled. The disentanglement ability of our model comes from its accurate identification of the relevant latent variables subject to intervention in the causal graph when one factor of variation is modified. This difference, unfortunately, prevents us from comparing the level of disentanglement of our model using standard metrics like DCI \cite{DBLP:conf/iclr/EastwoodW18} or SAP \cite{DBLP:conf/iclr/0001SB18}. We leave the question of developing precise disentanglement measures for quantized latent spaces for future work.

\bibliographystyle{named}
\bibliography{references/related_work, references/causal_latent_discovery, references/experiments, references/introduction, references/discussion}

\clearpage

\appendix

\section{Architecture details}

\subsection{MCQ-VAE details}

\begin{figure*}[t]
    \centering
    \begin{subfigure}{0.45\textwidth}
        \centering
        \includegraphics[width=\linewidth]{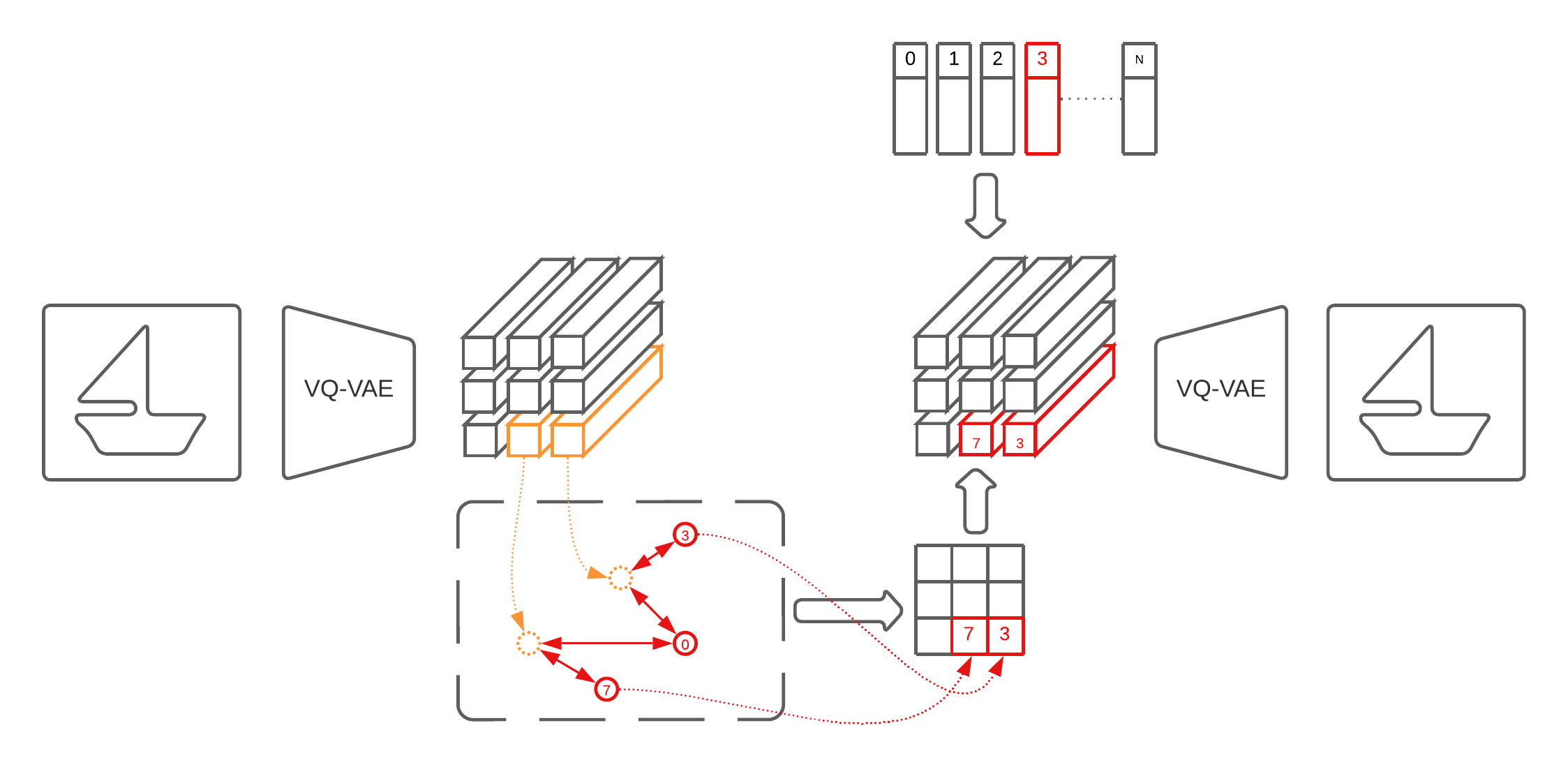}
        \caption{VQ-VAE. }
    \end{subfigure}
    \begin{subfigure}{0.45\textwidth}
        \centering
        \includegraphics[width=\linewidth]{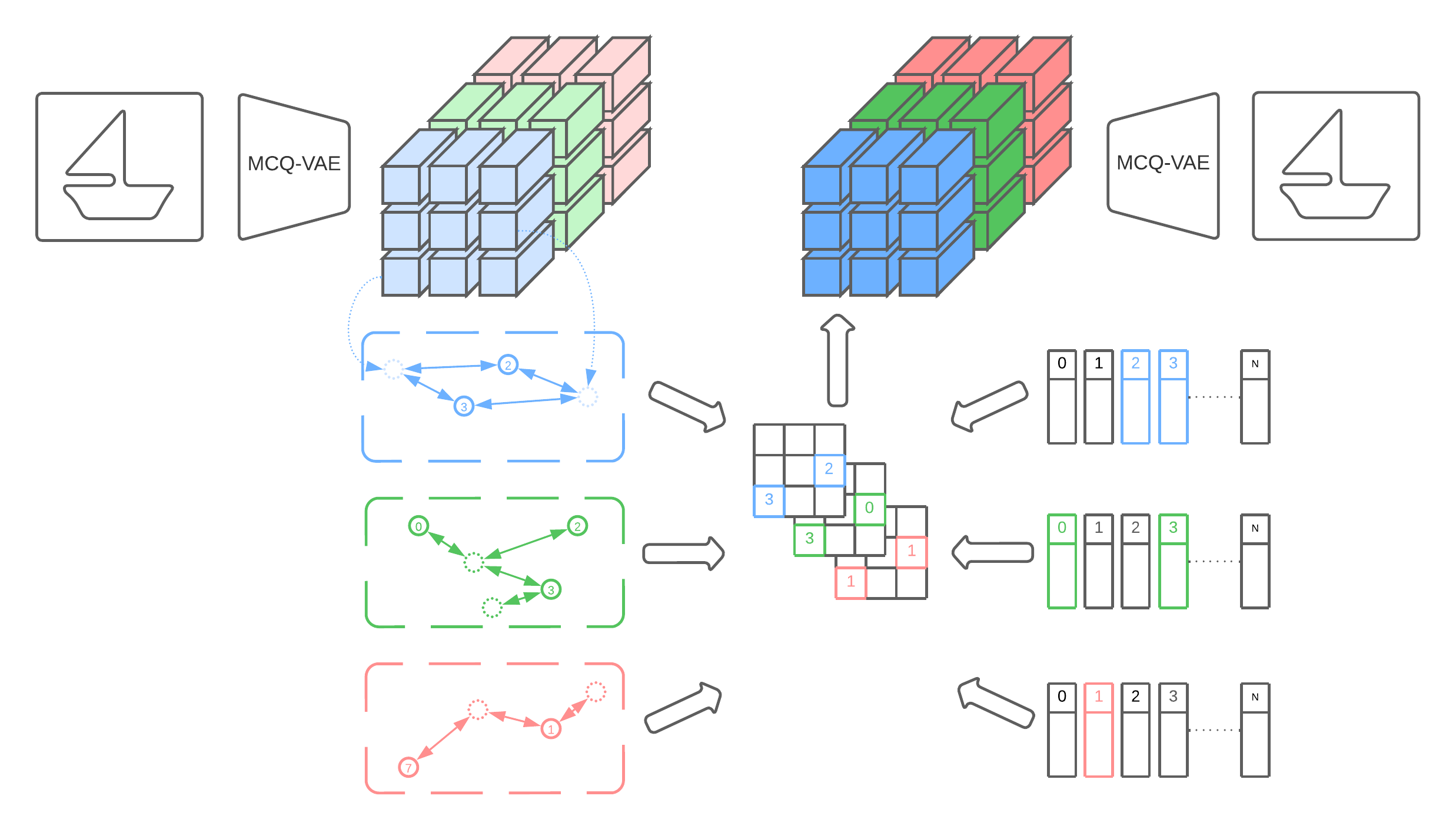}
        \caption{MCQ-VAE. }
    \end{subfigure}
    \caption{Overview and comparison of the VQ-VAE and MCQ-VAE architectures. }
    \label{fig:vq-mcq-vae_comp}
\end{figure*}

The \textit{Multi-Codebook Quantized VAE} or MCQ-VAE is an architecture based on the VQ-VAE \cite{DBLP:conf/nips/OordVK17} using multiple codebooks instead of a single one. Figure \ref{fig:vq-mcq-vae_comp} illustrates this difference graphically. During the pre-training stage, the MCQ-VAE is trained without the Causal Transition layer. In our experiments, we use the standard implementation of the VQ-VAE for our encoder and decoder. By default, we use a single codebook. The inputs consist of $64 \times 64$ images, which produce a latent space of size $8 \times 8$. The embedding dimension is $128$, with $64$ quantized embedding vectors. We set the weight of the embedding loss to $0.25$.

\subsection{Implementation details}

In our implementation, we use cross-entropy loss functions to compute the likelihoods $\mathbb{E}_{\mathbf{L_x}}$, $\mathbb{E}_{\mathbf{L_y}}$, and $\mathbb{E}_{\mathbf{a}}$. The Graph Neural Network comprises 3 Graph Attention (GAT) layers \cite{DBLP:conf/iclr/VelickovicCCRLB18}, using the static attention fix from GATv2 \cite{DBLP:conf/iclr/Brody0Y22} with ReLU activation. The output layer is a linear layer with softmax activation. We use a dense layer to project the action $\mathbf{a}$ to the same dimensions as $\mathbf{L_x}$ before feeding it to the GNN. The Multi-Layer Perceptrons used for causal structure discovery have two layers with ReLU and sigmoid activations. The binary intervention mask is obtained via Bernoulli trials with each parameter learned by backpropagation. The Straight-Through Gumbel-Softmax reparametrisation trick is used to allow gradient flow as in the trials for $\mathbf{M^\mathcal{G}}$. We add a 1-dimensional positional encoding to $\mathbf{L_x}$ in the input of the MLP and GNN computations, as in \cite{DBLP:conf/iclr/DosovitskiyB0WZ21}.

During the fine-tuning stage, the parameters of the autoencoder are frozen, and only the CT layer is trained. We reduce the weight of the embedding loss of the quantisation layer to $0.1$. We give a weight to each of the losses of the CT layer as follows:

\begin{equation}
    \begin{aligned}
        \mathcal{L} &= \gamma \cdot (\mathcal{L}_x + \mathcal{L}_y + \mathcal{L}_a \\
        &+ \alpha \cdot (\mathcal{L}_{id_y} + \mathcal{L}_{id_{M^\mathcal{G}}}) \\
        &+ \beta \cdot \text{KL} 
        + \delta \cdot \mathcal{L}_{|M^{\mathcal{G}}|} 
        + \epsilon \cdot \mathcal{L}_{dep(M^{\mathcal{G}})} ) 
    \end{aligned}
\end{equation}

$\gamma$ is a factor balancing the effect of all CT losses against VQ and reconstruction losses. $\alpha$ controls the importance of the identity losses, $\beta$, $\delta$, and $\epsilon$ control the regularisation of the causal graph. The values of $\gamma$, $\alpha$, $\beta$, $\delta$, and $\epsilon$ are set to $1.5$, $0.01$, $0.4$, $0.01$ and $0.1$, respectively.

\section{Additional Experiments}

\subsection{Results from additional datasets}

\begin{figure}[t]
    \centering
    \scriptsize
    \newcommand\nb{4}
    \newcommand\nnb{5}
    \newcommand{\datasetname}{TCars3D}
    \newcommand{\testfolder}{test1}
    \begin{tabular}{cccccc}
         & Input 
        \forloop{actionnum}{1}{\value{actionnum} < \nnb}{ & Output \arabic{actionnum}} \\
        \begin{sideways} Elevation \end{sideways} & \includegraphics[width=0.12\linewidth,trim=5 5 5 5,clip]{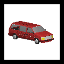}
        \forloop{actionnum}{0}{\value{actionnum} < \nb}{
            & \includegraphics[width=0.12\linewidth,trim=5 5 5 5,clip]{results/appendix/\datasetname_\testfolder/CT-VAE-V5_output_action_\datasetname_elevation_+\arabic{actionnum}_\testfolder.png}
        } \\
        \begin{sideways} Azimuth \end{sideways} & \includegraphics[width=0.12\linewidth,trim=5 5 5 5,clip]{results/appendix/\datasetname_\testfolder/CT-VAE-V5_input_action_\datasetname_\testfolder.png}
        \forloop{actionnum}{0}{\value{actionnum} < \nb}{
            & \includegraphics[width=0.12\linewidth,trim=5 5 5 5,clip]{results/appendix/\datasetname_\testfolder/CT-VAE-V5_output_action_\datasetname_azimuth_-\arabic{actionnum}_\testfolder.png}
        } \\
        \begin{sideways} Object Type \end{sideways} & \includegraphics[width=0.12\linewidth,trim=5 5 5 5,clip]{results/appendix/\datasetname_\testfolder/CT-VAE-V5_input_action_\datasetname_\testfolder.png}
        \forloop{actionnum}{0}{\value{actionnum} < \nb}{
            & \includegraphics[width=0.12\linewidth,trim=5 5 5 5,clip]{results/appendix/\datasetname_\testfolder/CT-VAE-V5_output_action_\datasetname_object_type_+\arabic{actionnum}_\testfolder.png}
        } \\
    \end{tabular}
    \caption{Atomic interventions on one factor of variation on images from the Cars3D dataset. Each row corresponds to an intervention on a different factor, with the first row being the input image. Output $i$ corresponds to the output after applying the same action $i$ times. }
    \label{fig:images_cars3d}
\end{figure}

\paragraph{Cars3D} We reconstruct images from the Cars3D dataset \cite{DBLP:conf/nips/ReedZZL15} with and without action application. Examples are shown in Figure \ref{fig:images_cars3d}. The quality of the generation is lower than for the other datasets. The Cars3D dataset contains only 3 factors of variation: the type of the car, camera elevation, and azimuth, with dimension sizes of 4, 24, and 185, respectively. A single factor controls the shape and colour of the car. We believe this is not sufficient to learn a good disentangled representation. Further experiments detailed in the following paragraphs show that our method works better with small changes, which do not match the labelling of the Cars3D dataset.

\begin{figure}[t]
    \centering
    \scriptsize
    \newcommand\nb{4}
    \newcommand\nnb{5}
    \newcommand{\datasetname}{TSprites}
    \newcommand{\testfolder}{test1}
    \begin{tabular}{cccccc}
         & Input 
        \forloop{actionnum}{1}{\value{actionnum} < \nnb}{ & Output \arabic{actionnum}} \\
        \begin{sideways} Bottomwear \end{sideways} & \includegraphics[width=0.12\linewidth,trim=20 20 20 20,clip]{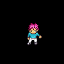}
        \forloop{actionnum}{0}{\value{actionnum} < \nb}{
            & \includegraphics[width=0.12\linewidth,trim=20 20 20 20,clip]{results/appendix/\datasetname_\testfolder/CT-VAE-V5_output_action_\datasetname_Bottomwear_+\arabic{actionnum}_\testfolder.png}
        } \\
        \begin{sideways} Topwear \end{sideways} & \includegraphics[width=0.12\linewidth,trim=20 20 20 20,clip]{results/appendix/\datasetname_\testfolder/CT-VAE-V5_input_action_\datasetname_\testfolder.png}
        \forloop{actionnum}{0}{\value{actionnum} < \nb}{
            & \includegraphics[width=0.12\linewidth,trim=20 20 20 20,clip]{results/appendix/\datasetname_\testfolder/CT-VAE-V5_output_action_\datasetname_topwear_-\arabic{actionnum}_\testfolder.png}
        } \\
        \begin{sideways} Hair \end{sideways} & \includegraphics[width=0.12\linewidth,trim=20 20 20 20,clip]{results/appendix/\datasetname_\testfolder/CT-VAE-V5_input_action_\datasetname_\testfolder.png}
        \forloop{actionnum}{0}{\value{actionnum} < \nb}{
            & \includegraphics[width=0.12\linewidth,trim=20 20 20 20,clip]{results/appendix/\datasetname_\testfolder/CT-VAE-V5_output_action_\datasetname_hair_-\arabic{actionnum}_\testfolder.png}
        } \\
        \begin{sideways} Eyes \end{sideways} & \includegraphics[width=0.12\linewidth,trim=20 20 20 20,clip]{results/appendix/\datasetname_\testfolder/CT-VAE-V5_input_action_\datasetname_\testfolder.png}
        \forloop{actionnum}{0}{\value{actionnum} < \nb}{
            & \includegraphics[width=0.12\linewidth,trim=20 20 20 20,clip]{results/appendix/\datasetname_\testfolder/CT-VAE-V5_output_action_\datasetname_eyes_-\arabic{actionnum}_\testfolder.png}
        } \\
        \begin{sideways} Shoes \end{sideways} & \includegraphics[width=0.12\linewidth,trim=20 20 20 20,clip]{results/appendix/\datasetname_\testfolder/CT-VAE-V5_input_action_\datasetname_\testfolder.png}
        \forloop{actionnum}{0}{\value{actionnum} < \nb}{
            & \includegraphics[width=0.12\linewidth,trim=20 20 20 20,clip]{results/appendix/\datasetname_\testfolder/CT-VAE-V5_output_action_\datasetname_shoes_-\arabic{actionnum}_\testfolder.png}
        } \\
        \begin{sideways} Body \end{sideways} & \includegraphics[width=0.12\linewidth,trim=20 20 20 20,clip]{results/appendix/\datasetname_\testfolder/CT-VAE-V5_input_action_\datasetname_\testfolder.png}
        \forloop{actionnum}{0}{\value{actionnum} < \nb}{
            & \includegraphics[width=0.12\linewidth,trim=20 20 20 20,clip]{results/appendix/\datasetname_\testfolder/CT-VAE-V5_output_action_\datasetname_body_+\arabic{actionnum}_\testfolder.png}
        } \\
        \begin{sideways} Action/Pose \end{sideways} & \includegraphics[width=0.12\linewidth,trim=20 20 20 20,clip]{results/appendix/\datasetname_\testfolder/CT-VAE-V5_input_action_\datasetname_\testfolder.png}
        \forloop{actionnum}{0}{\value{actionnum} < \nb}{
            & \includegraphics[width=0.12\linewidth,trim=20 20 20 20,clip]{results/appendix/\datasetname_\testfolder/CT-VAE-V5_output_action_\datasetname_action_-\arabic{actionnum}_\testfolder.png}
        } \\
        \begin{sideways} Rotation \end{sideways} & \includegraphics[width=0.12\linewidth,trim=20 20 20 20,clip]{results/appendix/\datasetname_\testfolder/CT-VAE-V5_input_action_\datasetname_\testfolder.png}
        \forloop{actionnum}{0}{\value{actionnum} < \nb}{
            & \includegraphics[width=0.12\linewidth,trim=20 20 20 20,clip]{results/appendix/\datasetname_\testfolder/CT-VAE-V5_output_action_\datasetname_rotation_-\arabic{actionnum}_\testfolder.png}
        } \\
        \begin{sideways} Frame \end{sideways} & \includegraphics[width=0.12\linewidth,trim=20 20 20 20,clip]{results/appendix/\datasetname_\testfolder/CT-VAE-V5_input_action_\datasetname_\testfolder.png}
        \forloop{actionnum}{0}{\value{actionnum} < \nb}{
            & \includegraphics[width=0.12\linewidth,trim=20 20 20 20,clip]{results/appendix/\datasetname_\testfolder/CT-VAE-V5_output_action_\datasetname_frame_-\arabic{actionnum}_\testfolder.png}
        } \\
    \end{tabular}
    \caption{Atomic interventions on one factor of variation on images from the Sprites dataset. Each row corresponds to an intervention on a different factor, with the first row being the input image. Output $i$ corresponds to the output after applying the same action $i$ times. }
    \label{fig:images_sprites}
\end{figure}

\paragraph{Sprites} We reconstruct some images from the Sprites dataset \cite{DBLP:conf/nips/ReedZZL15} with and without action application. Examples are shown in Figure \ref{fig:images_sprites}. We can observe that the quality of the generated images suffers little from the application of the action. In particular, modifying colour attributes such as hair colour, topwear, or bottomwear, does not affect the quality of the generation. Only the intervened factor of variation is modified. On the other hand, interventions yielding a more significant change in the image, like pose or rotation, are harder to achieve.

\begin{figure}[t]
    \centering
    \scriptsize
    \newcommand\nb{4}
    \newcommand\nnb{5}
    \newcommand{\datasetname}{TDSprites}
    \newcommand{\testfolder}{test1}
    \begin{tabular}{cccccc}
         & Input 
        \forloop{actionnum}{1}{\value{actionnum} < \nnb}{ & Output \arabic{actionnum}} \\
        \begin{sideways} Shape \end{sideways} & \includegraphics[width=0.12\linewidth,trim=20 20 20 20,clip]{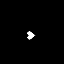}
        \forloop{actionnum}{0}{\value{actionnum} < \nb}{
            & \includegraphics[width=0.12\linewidth,trim=20 20 20 20,clip]{results/appendix/\datasetname_\testfolder/CT-VAE-V5_output_action_\datasetname_shape_-\arabic{actionnum}_\testfolder.png}
        } \\
        \begin{sideways} Scale \end{sideways} & \includegraphics[width=0.12\linewidth,trim=20 20 20 20,clip]{results/appendix/\datasetname_\testfolder/CT-VAE-V5_input_action_\datasetname_\testfolder.png}
        \forloop{actionnum}{0}{\value{actionnum} < \nb}{
            & \includegraphics[width=0.12\linewidth,trim=20 20 20 20,clip]{results/appendix/\datasetname_\testfolder/CT-VAE-V5_output_action_\datasetname_scale_-\arabic{actionnum}_\testfolder.png}
        } \\
        \begin{sideways} Orientation \end{sideways} & \includegraphics[width=0.12\linewidth,trim=20 20 20 20,clip]{results/appendix/\datasetname_\testfolder/CT-VAE-V5_input_action_\datasetname_\testfolder.png}
        \forloop{actionnum}{0}{\value{actionnum} < \nb}{
            & \includegraphics[width=0.12\linewidth,trim=20 20 20 20,clip]{results/appendix/\datasetname_\testfolder/CT-VAE-V5_output_action_\datasetname_orientation_-\arabic{actionnum}_\testfolder.png}
        } \\
        \begin{sideways} Position X \end{sideways} & \includegraphics[width=0.12\linewidth,trim=20 20 20 20,clip]{results/appendix/\datasetname_\testfolder/CT-VAE-V5_input_action_\datasetname_\testfolder.png}
        \forloop{actionnum}{0}{\value{actionnum} < \nb}{
            & \includegraphics[width=0.12\linewidth,trim=20 20 20 20,clip]{results/appendix/\datasetname_\testfolder/CT-VAE-V5_output_action_\datasetname_position_x_-\arabic{actionnum}_\testfolder.png}
        } \\
        \begin{sideways} Position Y \end{sideways} & \includegraphics[width=0.12\linewidth,trim=20 20 20 20,clip]{results/appendix/\datasetname_\testfolder/CT-VAE-V5_input_action_\datasetname_\testfolder.png}
        \forloop{actionnum}{0}{\value{actionnum} < \nb}{
            & \includegraphics[width=0.12\linewidth,trim=20 20 20 20,clip]{results/appendix/\datasetname_\testfolder/CT-VAE-V5_output_action_\datasetname_position_y_-\arabic{actionnum}_\testfolder.png}
        } \\
    \end{tabular}
    \caption{Atomic interventions on one factor of variation on images from the DSprites dataset. Each row corresponds to an intervention on a different factor, with the first row being the input image. Output $i$ corresponds to the output after applying the same action $i$ times. }
    \label{fig:images_dsprites}
\end{figure}

\paragraph{DSprites and Shapes3D} We reconstruct some images from the DSprites dataset \cite{DBLP:conf/iclr/HigginsMPBGBML17} with and without action application. Examples are shown in Figure \ref{fig:images_dsprites}.

We look more closely at the causal accuracy for the DSprites dataset. Table \ref{tab:causal_results_dsprites} summarises the results. The orientation action is the easiest to detect for our model, while it struggles to recover the others. In particular, it consistently fails to recognise a change in the object scale. 
Table \ref{tab:causal_results_shapes3d} shows the causal accuracy for the Shapes3D dataset \cite{DBLP:conf/icml/KimM18}. The results are much better, which is consistent with the observed quality of the generation. By comparing the two tables, we can see that the actions involving changes on a different set of variables between $\mathbf{L_x}$ and $\mathbf{L_y}$ are the hardest to recognise for the model. These actions comprise the position, shape, or scale on the DSprites dataset or the shape and scale on the Shapes3D dataset.

\begin{table*}[t]
    \centering
    \begin{tabular}{lrrrrrrrrrr}
        \hline
          Actions & \multicolumn{2}{c}{Shape} & \multicolumn{2}{c}{Scale} & \multicolumn{2}{c}{Orientation} & \multicolumn{2}{c}{Position X} & \multicolumn{2}{c}{Position Y} \\
        Direction & - & + & - & + & - & + & - & + & - & + \\
        \hline
        Cardinality & 3 & 3 & 6 & 6 & 40 & 40 & 32 & 32 & 32 & 32 \\
        Action Acc. & 0.08 & 0.20 & 0.00 & 0.00 & 0.48 & 0.69 & 0.22 & 0.33 & 0.25 & 0.38 \\
        Factor Acc. & 0.21 & 0.23 & 0.00 & 0.00 & 0.89 & 0.98 & 0.22 & 0.33 & 0.28 & 0.38 \\
        \hline
    \end{tabular}
    \caption{Accuracy of \textit{causal} mode for each action of the DSprites dataset. Each factor of variation has two associated actions, one increasing (+) and the other decreasing (-) its value. \textit{Action Acc.} shows the action accuracy, and \textit{Factor Acc.} shows the factor accuracy. The results show the average over a batch of 64 input-output couples of images. }
    \label{tab:causal_results_dsprites}
\end{table*}

\begin{table*}[t]
    \centering
    \begin{tabular}{lrrrrrrrrrrrr}
        \hline
          Actions & \multicolumn{2}{c}{Floor hue} & \multicolumn{2}{c}{Wall hue} & \multicolumn{2}{c}{Object hue} & \multicolumn{2}{c}{Scale} & \multicolumn{2}{c}{Shape} & \multicolumn{2}{c}{Orientation} \\
        Direction & - & + & - & + & - & + & - & + & - & + & - & + \\
        \hline
        Cardinality & 10 & 10 & 10 & 10 & 10 & 10 & 8 & 8 & 4 & 4 & 15 & 15 \\
        Action Acc. & 0.73 & 1.00 & 1.00 & 0.84 & 0.84 & 0.88 & 0.36 & 0.50 & 0.13 & 0.22 & 0.52 &0.47 \\
        Factor Acc. & 0.73 & 1.00 & 1.00 & 0.91 & 0.88 & 0.91 & 0.45 & 0.51 & 0.13 & 0.22 & 0.73 & 0.70 \\
        \hline
    \end{tabular}
    \caption{Accuracy of \textit{causal} mode for each action of the Shapes3D dataset. Each factor of variation has two associated actions, one increasing (+) and the other decreasing (-) its value. \textit{Action Acc.} shows the action accuracy, and \textit{Factor Acc.} shows the factor accuracy. The results show the average over a batch of 64 input-output couples of images. }
    \label{tab:causal_results_shapes3d}
\end{table*}

\begin{figure*}[t]
    \centering
    \begin{subfigure}{0.19\textwidth}
        \centering
        \includegraphics[width=\linewidth]{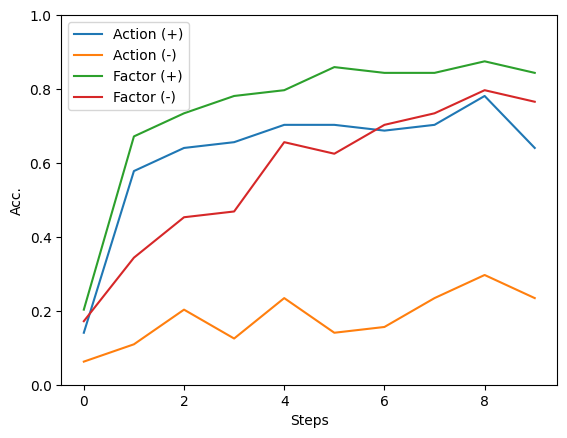}
    \caption{Shape. }
    \end{subfigure}
    \hfill
    \begin{subfigure}{0.19\textwidth}
        \centering
        \includegraphics[width=\linewidth]{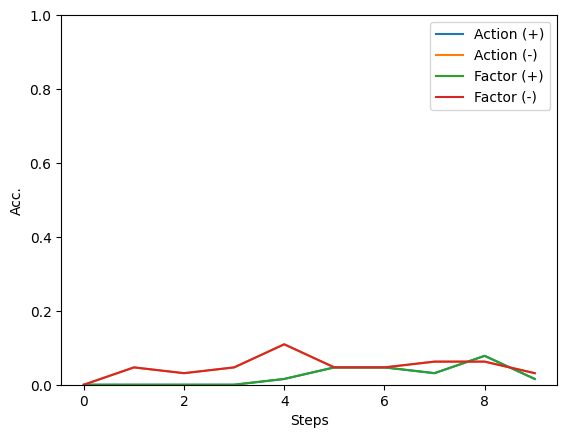}
    \caption{Scale. }
    \end{subfigure}
    \hfill
    \begin{subfigure}{0.19\textwidth}
        \centering
        \includegraphics[width=\linewidth]{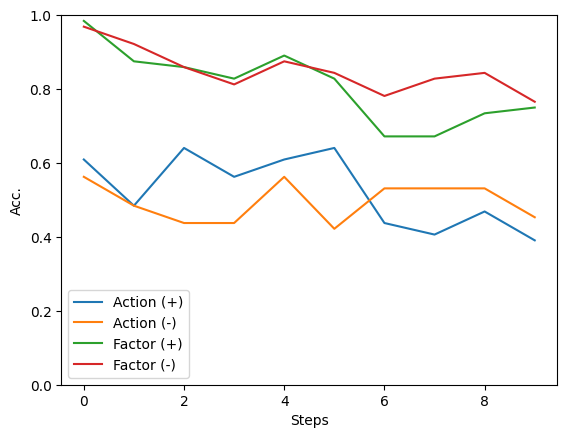}
    \caption{Orientation. }
    \end{subfigure}
    \hfill
    \begin{subfigure}{0.19\textwidth}
        \centering
        \includegraphics[width=\linewidth]{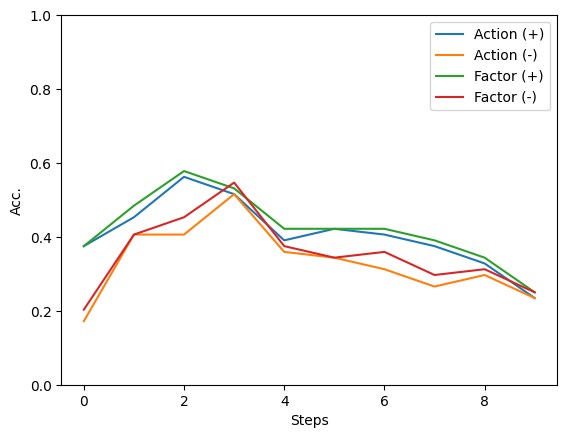}
    \caption{Position X. }
    \end{subfigure}
    \hfill
    \begin{subfigure}{0.19\textwidth}
        \centering
        \includegraphics[width=\linewidth]{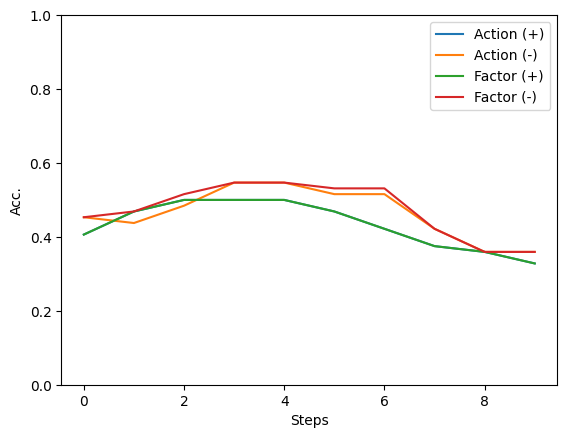}
    \caption{Position Y. }
    \end{subfigure}
    \caption{Evolution of the causal accuracy when repeating an action for multiple steps on the DSprites dataset. Each plot shows the action and factor accuracies for the repeated action in the two directions of the axis. The results show the average over 64 input-output couples of images. }
    \label{fig:dsprites_acc_plot}
\end{figure*}

\begin{figure*}[t]
    \centering
    \begin{subfigure}{0.16\textwidth}
        \centering
        \includegraphics[width=\linewidth]{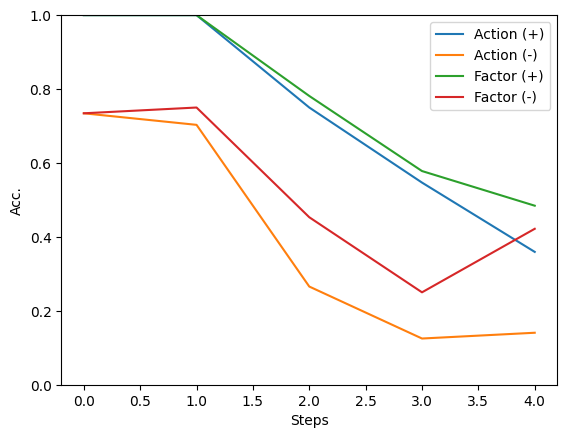}
    \caption{Floor hue. }
    \end{subfigure}
    \hfill
    \begin{subfigure}{0.16\textwidth}
        \centering
        \includegraphics[width=\linewidth]{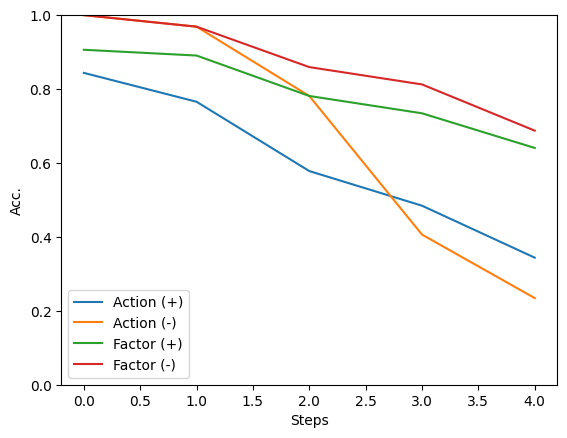}
    \caption{Wall hue. }
    \end{subfigure}
    \hfill
    \begin{subfigure}{0.16\textwidth}
        \centering
        \includegraphics[width=\linewidth]{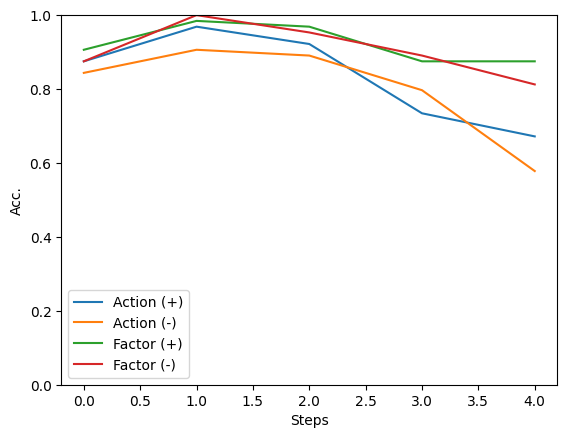}
    \caption{Object hue. }
    \end{subfigure}
    \hfill
    \begin{subfigure}{0.16\textwidth}
        \centering
        \includegraphics[width=\linewidth]{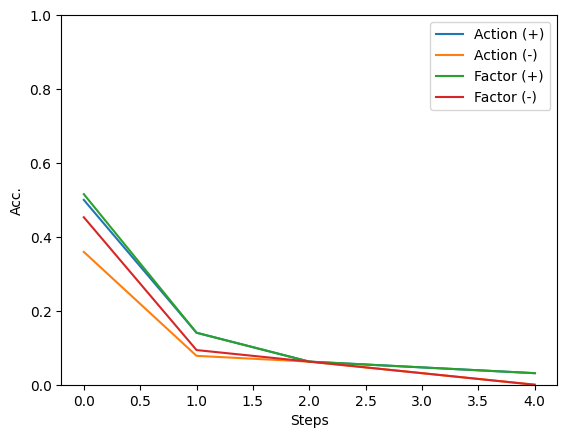}
    \caption{Scale. }
    \end{subfigure}
    \hfill
    \begin{subfigure}{0.16\textwidth}
        \centering
        \includegraphics[width=\linewidth]{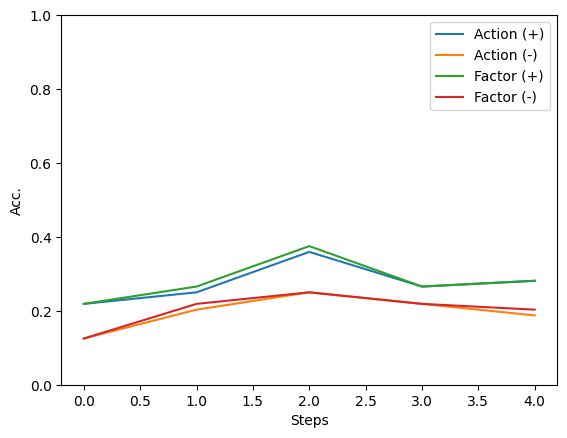}
    \caption{Shape. }
    \end{subfigure}
    \hfill
    \begin{subfigure}{0.16\textwidth}
        \centering
        \includegraphics[width=\linewidth]{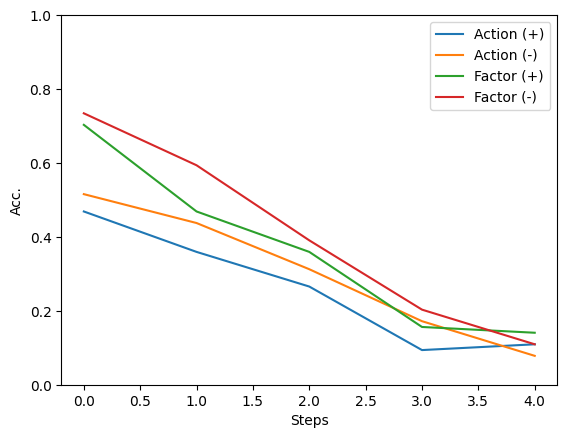}
    \caption{Orientation. }
    \end{subfigure}
    \caption{Evolution of the causal accuracy when repeating an action for multiple steps on the Shapes3D dataset. Each plot shows the action and factor accuracies for the repeated action in the two directions of the axis. The results show the average over 64 input-output couples of images. }
    \label{fig:shapes3d_acc_plot}
\end{figure*}

We perform additional experiments on DSprites and Shapes3D to study the impact of significant changes in the image on causal accuracy. We perform a sequence of identical actions on a batch of images and compare the reconstructed images with the original batch at each step. The sequence repeats 8 times the action for the DSprites dataset and 4 times for the Shapes3D dataset. The resulting causal accuracies for each action are represented in Figures \ref{fig:dsprites_acc_plot} and \ref{fig:shapes3d_acc_plot}. The accuracy for the actions based on colour change increases or remains constant when applying the first transitions, then decreases as the number of transitions increases further. This result tends to indicate that the model learns gradual changes. It can detect small changes in a factor of variation, e.g. floor hue evolving from red to orange, and starker changes help it to a degree. However, if the change is too big, the model will struggle to recognise the transition involved. The same can be observed for the object position. Curiously, the object shape in the DSprites dataset is the factor getting the most benefits by adding actions. This benefit is not observed in the Shapes3D dataset.

\subsection{Results in the presence of confounders}

\begin{figure*}[t]
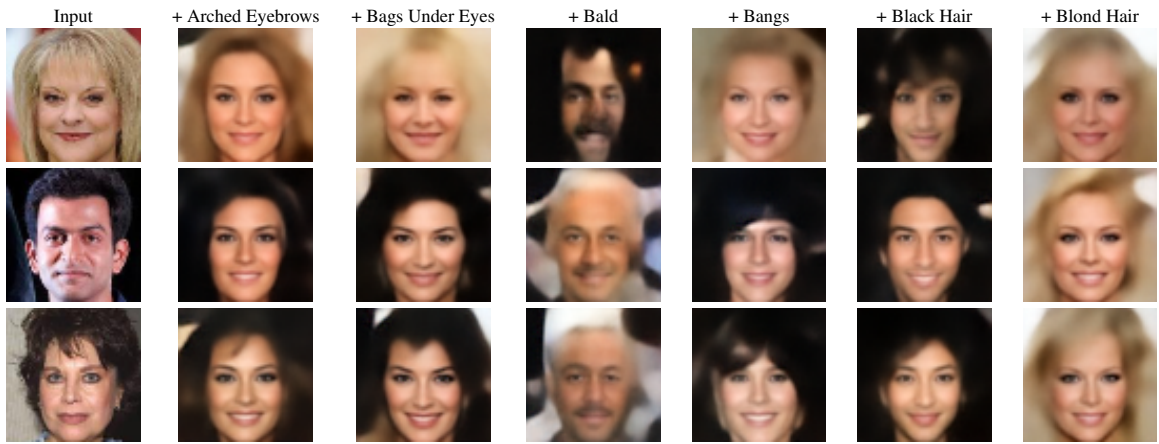

    \centering
    \scriptsize
    \newcounter{testnum}
    \newcommand{\datasetname}{TCeleba}
    \begin{tabular}{ccccccc}
         Input & + Arched Eyebrows & + Bags Under Eyes & + Bald & + Bangs & + Black Hair & + Blond Hair 
        \forloop{testnum}{1}{\value{testnum} < 4}{
            \\ \includegraphics[width=0.1\linewidth]{results/appendix/\datasetname_test\arabic{testnum}/CT-VAE-V5_input_action_\datasetname_test\arabic{testnum}.png} &
            \includegraphics[width=0.1\linewidth]{results/appendix/\datasetname_test\arabic{testnum}/CT-VAE-V5_output_action_\datasetname_Arched_Eyebrows_+0_test\arabic{testnum}.png} & \includegraphics[width=0.1\linewidth]{results/appendix/\datasetname_test\arabic{testnum}/CT-VAE-V5_output_action_\datasetname_Bags_Under_Eyes_+0_test\arabic{testnum}.png} & \includegraphics[width=0.1\linewidth]{results/appendix/\datasetname_test\arabic{testnum}/CT-VAE-V5_output_action_\datasetname_Bald_+0_test\arabic{testnum}.png} & \includegraphics[width=0.1\linewidth]{results/appendix/\datasetname_test\arabic{testnum}/CT-VAE-V5_output_action_\datasetname_Bangs_+0_test\arabic{testnum}.png} & \includegraphics[width=0.1\linewidth]{results/appendix/\datasetname_test\arabic{testnum}/CT-VAE-V5_output_action_\datasetname_Black_Hair_+0_test\arabic{testnum}.png} & \includegraphics[width=0.1\linewidth]{results/appendix/\datasetname_test\arabic{testnum}/CT-VAE-V5_output_action_\datasetname_Blond_Hair_+0_test\arabic{testnum}.png} 
        }
    \end{tabular}
    \caption{Atomic interventions on one factor of variation on images from the Celeba dataset. Each column corresponds to an intervention on a different factor, with the first column being the input image. (+) means the the attribute is added to the image. }
    \label{fig:images_celeba}
\end{figure*}

The CelebA dataset \cite{DBLP:conf/iccv/LiuLWT15} is a real-world dataset of celebrity faces. Unlike the above datasets, the labelled attributes do not fully characterise each image. These missing attributes correspond to exogenous factors of variation. We attempt to represent these variables as noise added to the causal graph. We focus our work on a subset of CelebA with 6 binary attributes: "arched eyebrows", "bags under eyes", "bald", "bangs", "black hair", and "blond hair". Figure \ref{fig:images_celeba} illustrates the results obtained with this dataset. Our model can generate realistic images when no action is applied but struggle in action mode. When performing an action, the reconstructed image contains the modified attribute as expected, but other factors of variation are also modified. For instance, when changing the hair colour to blonde, the face often becomes the face of a woman. Our model is fooled by spurious associations. In this instance, the blonde hair colour is associated with the gender variable. Similarly, the bald attribute is associated with men. When intervening on this factor from images of women, the quality of the generated results decreases notably.

We study the causal accuracy of the 6 labelled factors from the subset of CelebA used in our work. The results are shown in Table \ref{tab:causal_results_celeba}. The accuracy is very different depending on the action applied and the direction of the action. This behaviour differs from the other datasets and can be explained by the binary nature of the factors of variation, as opposed to the categorical attributes of the synthetic datasets. The most extreme representative is the bald attribute. When removing this attribute, the accuracy is very low, but it is close to one when adding it.
Unlike the other datasets, the factor accuracy is not better than the action accuracy. It significates that when the model succeeds in recovering the factor of variation, it can also identify if the attribute is being added or removed.

\begin{table*}[t]
    \centering
    \begin{tabular}{lrrrrrrrrrrrr}
        \hline
          Actions & \multicolumn{2}{c}{Arched Eyebrows} & \multicolumn{2}{c}{Bags Under Eyes} & \multicolumn{2}{c}{Bald} & \multicolumn{2}{c}{Bangs} & \multicolumn{2}{c}{Black Hair} & \multicolumn{2}{c}{Blond Hair} \\
        Direction & - & + & - & + & - & + & - & + & - & + & - & + \\
        \hline
        Action Acc. & 0.95 & 1.0 & 0.98 & 0.88 & 0.30 & 0.98 & 0.78 & 0.86 & 0.90 & 1.0 & 1.0 & 1.0 \\
        Factor Acc. & 0.95 & 1.0 & 0.98 & 0.88 & 0.30 & 0.98 & 0.78 & 0.86 & 0.94 & 1.0 & 1.0 & 1.0 \\
        \hline
    \end{tabular}
    \caption{Accuracy of \textit{causal} mode for each action of the CelebA dataset. Each factor of variation has two associated actions, one increasing (+) and the other decreasing (-) its value. All attributes are binary and therefore have a cardinality of 2. \textit{Action Acc.} shows the action accuracy, and \textit{Factor Acc.} shows the factor accuracy. The results show the average over a batch of 64 input-output couples of images. }
    \label{tab:causal_results_celeba}
\end{table*}

\subsection{Ablation studies}

\begin{table}[t]
    \centering
    \begin{tabular}{llrr}
        \hline
         & & Action Acc. & Factor Acc. \\
        \hline
        & Base & 0.85 & 0.90 \\
        (+) & Endogenous noise & \textit{0.86} & 0.91 \\
        (+) & Exogenous noise & 0.71 & 0.84 \\
        (+) & 2 codebooks & 0.72 & \textit{0.94} \\
        (+) & 4 codebooks & \textbf{0.87} & \textbf{0.95} \\
        (-) & 1-layer GNN & 0.55 & 0.79 \\
        (-) & No mask & 0.08 & 0.17 \\
        \hline
    \end{tabular}
    \caption{Accuracy of \textit{causal} mode for the Shapes3D dataset on variations of the CT-VAE. The base model is used without noise, with 1 codebook and 3 GNN layers. The highest accuracies are written in \textbf{bold}, and the second best are in \textit{italics}. }
    \label{tab:causal_results_shapes3d_ablated}
\end{table}

\paragraph{Causal Masking}

We create an ablated version of the CT-VAE architecture without causal masking. Instead, the Bernoulli coefficients used to generate the causal graph are all obtained using the same network $MLP^{\emptyset}_{\phi}$. We study the causal accuracy with this ablated model. Results are displayed in Table \ref{tab:causal_results_shapes3d_ablated}. When the causal masking is removed, action and factor accuracy drop. For action accuracy, the model does not perform better than random. We deduce that causal masking is a crucial component for disentangling the factors of variation when performing causal transition.

\paragraph{Codebook}

We now take advantage of the ability of the MCQ-VAE architecture to use several codebooks and study the impact of several codebooks on causal accuracy. We use a model with the same number of parameters to allow fair comparison. Therefore, the embedding space of size $128$ is divided into two embedding vectors of size $64$. Table \ref{tab:causal_results_shapes3d_ablated} summarises the results obtained. Adding one codebook increases the factor accuracy by a thin margin but does not affect the action accuracy. The effect increases when using 4 codebooks. No differences were noticed in the quality of the generated images. However, adding a new codebook linearly increases the number of variables, which in turn causes a quadratic increase in the size of the causal graph, making the use of more than two codebooks intractable in practice.

\paragraph{GNN depth}

We reduce the depth of the Graph Neural Network to investigate the impact of multi-step node aggregation on the performance.  We can see in Table \ref{tab:causal_results_shapes3d_ablated} that reducing the depth of the GNN hurts the accuracy. This behaviour is consistent with the design of existing graph autoencoders for causal inference \cite{DBLP:journals/corr/abs-2109-04173,DBLP:conf/aaai/Sanchez-MartinR22}, and in particular, the design condition for the decoder of VACA \cite{DBLP:conf/aaai/Sanchez-MartinR22}, which states that the number of layers of the decoder should be greater or equal to the longest directed path in the causal graph.

\paragraph{Noise}

\begin{figure*}[t]
    \centering
    \scriptsize
    \newcommand\nb{3}
    \newcommand\nnb{4}
    \newcommand{\datasetname}{TShapes3D}
    \newcommand{\testfolder}{test3_noiseoff}
    \newcommand{\testfolderd}{test4_noiseendo}
    \newcommand{\testfolderdd}{test5_noiseexo}
    \begin{tabular}{cccccccccccccc}
         & & & \multicolumn{3}{c}{No noise} & & \multicolumn{3}{c}{Endogenous noise} & & \multicolumn{3}{c}{Exogenous noise} \\ \cline{4-6}\cline{8-10}\cline{12-14}
         & Input &
        \forloop{actionnum}{1}{\value{actionnum} < \nnb}{ & Output \arabic{actionnum}} &
        \forloop{actionnum}{1}{\value{actionnum} < \nnb}{ & Output \arabic{actionnum}} \\
        \begin{sideways} Floor hue \end{sideways} & \includegraphics[width=0.06\linewidth,trim=20 20 20 20,clip]{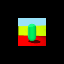} &
        \forloop{actionnum}{0}{\value{actionnum} < \nb}{
            & \includegraphics[width=0.06\linewidth,trim=20 20 20 20,clip]{results/appendix/\datasetname_\testfolder/CT-VAE-V5_output_action_\datasetname_floor_hue_+\arabic{actionnum}_\testfolder.png}
        } &
        \forloop{actionnum}{0}{\value{actionnum} < \nb}{
            & \includegraphics[width=0.06\linewidth,trim=20 20 20 20,clip]{results/appendix/\datasetname_\testfolderd/CT-VAE-V5_output_action_\datasetname_floor_hue_+\arabic{actionnum}_\testfolderd.png}
        } &
        \forloop{actionnum}{0}{\value{actionnum} < \nb}{
            & \includegraphics[width=0.06\linewidth,trim=20 20 20 20,clip]{results/appendix/\datasetname_\testfolderdd/CT-VAE-V5_output_action_\datasetname_floor_hue_+\arabic{actionnum}_\testfolderdd.png}
        } \\
        \begin{sideways} Wall hue \end{sideways} & \includegraphics[width=0.06\linewidth,trim=20 20 20 20,clip]{results/appendix/\datasetname_\testfolder/CT-VAE-V5_input_action_\datasetname_\testfolder.png} &
        \forloop{actionnum}{0}{\value{actionnum} < \nb}{
            & \includegraphics[width=0.06\linewidth,trim=20 20 20 20,clip]{results/appendix/\datasetname_\testfolder/CT-VAE-V5_output_action_\datasetname_wall_hue_+\arabic{actionnum}_\testfolder.png}
        } &
        \forloop{actionnum}{0}{\value{actionnum} < \nb}{
            & \includegraphics[width=0.06\linewidth,trim=20 20 20 20,clip]{results/appendix/\datasetname_\testfolderd/CT-VAE-V5_output_action_\datasetname_wall_hue_+\arabic{actionnum}_\testfolderd.png}
        } &
        \forloop{actionnum}{0}{\value{actionnum} < \nb}{
            & \includegraphics[width=0.06\linewidth,trim=20 20 20 20,clip]{results/appendix/\datasetname_\testfolderdd/CT-VAE-V5_output_action_\datasetname_wall_hue_+\arabic{actionnum}_\testfolderdd.png}
        } \\
        \begin{sideways} Object hue \end{sideways} & \includegraphics[width=0.06\linewidth,trim=20 20 20 20,clip]{results/appendix/\datasetname_\testfolder/CT-VAE-V5_input_action_\datasetname_\testfolder.png} &
        \forloop{actionnum}{0}{\value{actionnum} < \nb}{
            & \includegraphics[width=0.06\linewidth,trim=20 20 20 20,clip]{results/appendix/\datasetname_\testfolder/CT-VAE-V5_output_action_\datasetname_object_hue_+\arabic{actionnum}_\testfolder.png}
        } &
        \forloop{actionnum}{0}{\value{actionnum} < \nb}{
            & \includegraphics[width=0.06\linewidth,trim=20 20 20 20,clip]{results/appendix/\datasetname_\testfolderd/CT-VAE-V5_output_action_\datasetname_object_hue_+\arabic{actionnum}_\testfolderd.png}
        } &
        \forloop{actionnum}{0}{\value{actionnum} < \nb}{
            & \includegraphics[width=0.06\linewidth,trim=20 20 20 20,clip]{results/appendix/\datasetname_\testfolderdd/CT-VAE-V5_output_action_\datasetname_object_hue_+\arabic{actionnum}_\testfolderdd.png}
        } \\
        \begin{sideways} Scale \end{sideways} & \includegraphics[width=0.06\linewidth,trim=20 20 20 20,clip]{results/appendix/\datasetname_\testfolder/CT-VAE-V5_input_action_\datasetname_\testfolder.png} &
        \forloop{actionnum}{0}{\value{actionnum} < \nb}{
            & \includegraphics[width=0.06\linewidth,trim=20 20 20 20,clip]{results/appendix/\datasetname_\testfolder/CT-VAE-V5_output_action_\datasetname_scale_+\arabic{actionnum}_\testfolder.png}
        } &
        \forloop{actionnum}{0}{\value{actionnum} < \nb}{
            & \includegraphics[width=0.06\linewidth,trim=20 20 20 20,clip]{results/appendix/\datasetname_\testfolderd/CT-VAE-V5_output_action_\datasetname_scale_+\arabic{actionnum}_\testfolderd.png}
        } &
        \forloop{actionnum}{0}{\value{actionnum} < \nb}{
            & \includegraphics[width=0.06\linewidth,trim=20 20 20 20,clip]{results/appendix/\datasetname_\testfolderdd/CT-VAE-V5_output_action_\datasetname_scale_+\arabic{actionnum}_\testfolderdd.png}
        } \\
        \begin{sideways} Shape \end{sideways} & \includegraphics[width=0.06\linewidth,trim=20 20 20 20,clip]{results/appendix/\datasetname_\testfolder/CT-VAE-V5_input_action_\datasetname_\testfolder.png} &
        \forloop{actionnum}{0}{\value{actionnum} < \nb}{
            & \includegraphics[width=0.06\linewidth,trim=20 20 20 20,clip]{results/appendix/\datasetname_\testfolder/CT-VAE-V5_output_action_\datasetname_shape_-\arabic{actionnum}_\testfolder.png}
        } &
        \forloop{actionnum}{0}{\value{actionnum} < \nb}{
            & \includegraphics[width=0.06\linewidth,trim=20 20 20 20,clip]{results/appendix/\datasetname_\testfolderd/CT-VAE-V5_output_action_\datasetname_shape_+\arabic{actionnum}_\testfolderd.png}
        } &
        \forloop{actionnum}{0}{\value{actionnum} < \nb}{
            & \includegraphics[width=0.06\linewidth,trim=20 20 20 20,clip]{results/appendix/\datasetname_\testfolderdd/CT-VAE-V5_output_action_\datasetname_shape_+\arabic{actionnum}_\testfolderdd.png}
        } \\
        \begin{sideways} Orientation \end{sideways} & \includegraphics[width=0.06\linewidth,trim=20 20 20 20,clip]{results/appendix/\datasetname_\testfolder/CT-VAE-V5_input_action_\datasetname_\testfolder.png} &
        \forloop{actionnum}{0}{\value{actionnum} < \nb}{
            & \includegraphics[width=0.06\linewidth,trim=20 20 20 20,clip]{results/appendix/\datasetname_\testfolder/CT-VAE-V5_output_action_\datasetname_orientation_-\arabic{actionnum}_\testfolder.png}
        } &
        \forloop{actionnum}{0}{\value{actionnum} < \nb}{
            & \includegraphics[width=0.06\linewidth,trim=20 20 20 20,clip]{results/appendix/\datasetname_\testfolderd/CT-VAE-V5_output_action_\datasetname_orientation_+\arabic{actionnum}_\testfolderd.png}
        } &
        \forloop{actionnum}{0}{\value{actionnum} < \nb}{
            & \includegraphics[width=0.06\linewidth,trim=20 20 20 20,clip]{results/appendix/\datasetname_\testfolderdd/CT-VAE-V5_output_action_\datasetname_orientation_+\arabic{actionnum}_\testfolderdd.png}
        } \\
    \end{tabular}
    \caption{Atomic interventions on one factor of variation on images from the Shapes3D dataset. Each row corresponds to an intervention on a different factor, with the first row being the input image. Output $i$ corresponds to the output after applying the same action $i$ times. Results for three versions of the CT-VAE with endogenous, exogenous noise, or without noise are shown. }
    \label{fig:images_shapes3d_noise}
\end{figure*}

We compare the images generated with and without noise. Figure \ref{fig:images_shapes3d_noise} shows a single input on which we perform a series of actions with 3 variants of the CT-VAE. The \textit{no-noise} model does not add a noise variable $Z$ to the causal graph. The \textit{endogenous-noise} model adds a single noise variable to every variable $L_{xi}$. All variables share the same noise. It is considered another endogenous variable in the graph. The \textit{exogenous-noise} model adds Gaussian noise to every variable $L_{xi}$. The noise represents external processes applied to each variable independently. 

We observe in Figure \ref{fig:images_shapes3d_noise} a similar quality of representations between the \textit{no-noise} and \textit{exogenous-noise} models. The \textit{endogenous-noise} model slightly degrades the image quality, as can be observed for the wall hue variation. However, Table \ref{tab:causal_results_shapes3d_ablated} shows that adding exogenous noise reduces the causal accuracy both for the actions and the factors, while endogenous noise does not affect performance.

\end{document}